\documentclass{ieeeaccess}
\usepackage{cite}
\usepackage{amsmath,amssymb,amsfonts}
\usepackage{algorithmic}
\usepackage{graphicx}
\usepackage{textcomp}

\usepackage{setspace}
\usepackage[font={rm,footnotesize},
            labelfont={color=accessblue}
            ]{caption}
\usepackage{multirow}
\usepackage{comment}
\usepackage[normalem]{ulem}
\useunder{\uline}{\ul}{}
\usepackage{balance}
\usepackage{xcolor}

\def\BibTeX{{\rm B\kern-.05em{\sc i\kern-.025em b}\kern-.08em
    T\kern-.1667em\lower.7ex\hbox{E}\kern-.125emX}}
\begin{document}
\history{Date of publication: 23 October 2026. The final published version is available at http://doi.org/10.1109/ACCESS.2026.3746362}
\doi{10.1109/ACCESS.2026.3746362}

\title{SD-MVSum: Script-Driven Multimodal Video Summarization Method and Datasets}
\author{\uppercase{Manolis Mylonas, Charalampia Zerva, Evlampios Apostolidis and Vasileios Mezaris}
\IEEEmembership{Senior Member, IEEE}}
\address[]{Centre for Research and Technology Hellas (CERTH), Information Technologies Institute (ITI), 57001 Thessaloniki, Greece (e-mail: emylonas@iti.gr, charazerva@iti.gr, apostolid@iti.gr, bmezaris@iti.gr)}
\tfootnote{This work was supported in part by Project MediaPot, implemented in the framework of National Recovery and Resilience Plan Greece 2.0, funded by the European Union – NextGenerationEU, through the call RESEARCH-CREATE-INNOVATE under Grant TAEDK-06196; and in part by Project eXeLMM, implemented in the framework of H.F.R.I call “3rd Call for H.F.R.I.’s Research Projects to Support Faculty Members \& Researchers” under Grant 25957. The publication of the article in OA mode was financially supported by HEAL-Link.}

\corresp{Corresponding author: Vasileios Mezaris (e-mail: bmezaris@iti.gr).}

\begin{abstract}
In this work, we present a method and two large-scale datasets for Script-Driven Multimodal Video Summarization. The proposed method, SD-MVSum, builds on our earlier SD-VSum method for script-driven video summarization. SD-VSum considered the relevance of a provided script, which outlines the desired content of the summary, only with the visual content of the video. SD-MVSum takes into account, in addition to the visual modality, the relevance of the script with the spoken content (i.e., the audio transcript) of the video. The dependence between each considered pair of data modalities, i.e., script-video and script-transcript, is modeled using a new weighted cross-modal attention mechanism. This mechanism explicitly exploits the semantic similarity between the paired modalities in order to promote the parts of the full-length video with the highest relevance to the script. Furthermore, we extend two large-scale datasets for script-driven (S-VideoXum) and generic (MrHiSum) video summarization, to make them suitable for training and evaluation of script-driven multimodal video summarization methods. Experimental comparisons document the competitiveness of the proposed SD-MVSum method against other SotA approaches for script-driven and generic video summarization. Our new method and extended datasets are available at: https://github.com/IDT-ITI/SD-MVSum.
\end{abstract}


\titlepgskip=-15pt

\maketitle

\section{Introduction}
\label{sec:intro}
Various methods for text/query-driven video summarization have been proposed over the last years in the literature, aiming to assist the generation of summarized versions of a full-length video that are customized to the user's needs. In most cases, these needs are expressed using one or more keywords (e.g., ``changing tire'') \cite{10.1007/978-3-319-46484-8_1,8099712,10.5555/3540261.3541333} or a short sentence (e.g., ``a man is washing the car'') \cite{10.5555/3504035.3504062,10.1145/3372278.3390695}, since the relevant methods are not compatible with more extensive descriptions. Consequently, the generated video summaries exhibit limited visual and semantic diversity, as they mainly contain the video parts that match a short-form user query.

To tackle the aforementioned limitation of existing methods, in our previous work \cite{10.1145/3746027.3755821} we introduced the task of script-driven video summarization and released a relevant large-scale dataset (called S-VideoXum). Using this dataset, we trained a method (called SD-VSum) that gets as input a long-form script outlining the content of the desired video summary, and forms the summary by finding associations between the script and the visual content of the video based on a cross-modal attention mechanism. However, the spoken content in the video is also a rich source of information for spotting such associations. Driven by this observation, in this work we extend SD-VSum to leverage also the video's spoken content, forming the SD-MVSum method for script-driven multimodal video summarization. As part of this extension, we replace the standard cross-modal attention mechanism of SD-MVSum, by introducing a weighted cross-modal attention mechanism. The latter explicitly exploits the semantic similarity between a pair of data modalities when modeling their dependence, to promote the parts of the video with the highest relevance to the input script. Finally, to assist future research, we extend the previously released S-VideoXum dataset for script-driven video summarization \cite{10.1145/3746027.3755821}, as well as the large-scale MrHiSum dataset for video highlight detection and summarization \cite{10.5555/3666122.3667886}, making them suitable for the task of script-driven multimodal video summarization. Our main contributions are as follows:
\begin{itemize}
    \item We extend the SD-VSum method for script-driven video summarization, originally considering just the visual content of the video, to leverage also the video's spoken content (audio transcripts).
    \item We introduce a weighted cross-modal attention mechanism, which explicitly exploits the semantic similarity between a pair of data modalities when modeling their dependence, in order to promote the parts of the video with the highest relevance to the input script.
    \item We extend two large-scale datasets for script-driven (S-VideoXum) and generic (MrHiSum) video summarization, in order to make them suitable for training and evaluation of script-driven multimodal video summarization methods.
\end{itemize}

\section{Related Work}
\label{sec:literature}

\subsection{Text/Query-driven video summarization}

Early attempts on text- or query-driven video summarization were based on the use of probabilistic and submodular optimization frameworks. Sharghi et al. \cite{10.1007/978-3-319-46484-8_1,8099712} used probabilistic models to select video shots that were both important to the video and relevant to the query, while Vasudevan et al. \cite{10.1145/3123266.3123297} employed a submodular optimization framework to ensure the selected frames were relevant to the textual query, but also visually diverse, representative and aesthetically pleasing. A significant shift was observed with the emergence of deep learning. Wei et al. \cite{10.5555/3504035.3504062} introduced a semantic-attended network that learns to select representative video parts by minimizing the distance between generated summaries and human-provided descriptions. On a different basis, Zhang et al. \cite{Zhang2018QCVS} trained a query-conditioned Generative Adversarial Network (GAN) with a three-player loss, where the Generator aims to learn how to generate a summary based on a joint representation of the query and video, and the Discriminator tries to discriminate the ground-truth summary from the generated and a randomly-created one. 
More advanced approaches aimed to capture complex relationships. Jiang et al. \cite{10.1145/3323873.3325040} designed a hierarchical network with diverse cross-modal and self-attention mechanisms, to model query-related long-range temporal dependencies and take into account user-oriented diversity and stochastic factors. Xiao et al. \cite{Xiao_Zhao_Zhang_Yan_Yang_2020} used local self-attention and query-aware global attention to rank shots according to their semantic relationship with the user query, while Narasimhan et al. \cite{10.5555/3540261.3541333} introduced CLIP-It, a method using a multi-head language-guided attention mechanism to estimate frames' importance based on their visual relevance and their correlation with the user query. Towards addressing data scarcity, Xiao et al. \cite{9063637} pretrained a hierarchical self-attentive network for visual importance estimation on the ActivityNet Captions dataset \cite{8237345}, fine-tuned it using a reinforced caption generator, and developed a module that computes shot-level scores for a given query. Mujtaba et al. \cite{9902992} presented a query-driven approach that employs 2D CNNs and is designed to run on client devices aiming to provide tailored summaries based on individual user preferences. Huang et al. \cite{10222138} explored the use of self-supervision to generate pseudo-labels and model relationships between pseudo and human labels, and employed context-aware query representations to capture the relevance between visual and textual modalities. Guo et al. \cite{10889812} presented a method for query-driven multimodal video summarization that gets as input a short-form text, the video frames and the audio transcripts, and fuses them using a Transformer-based framework that applies both coarse-grained (taking all modalities into account) and fine-grained (combining text with video and audio individually) fusion, giving equal importance to each modality. Rodrigo et al. \cite{11298698} described a text-guided framework for automatic sports video summarization, that leverages contrastive language–image pretraining to classify video frames as highlight or non-highlight based on natural-language descriptions. Finally, in our previous work \cite{10.1145/3746027.3755821}, we extended the VideoXum large-scale dataset for video summarization by producing textual descriptions of the ground-truth summaries, and trained the SD-VSum method that aligns and fuses visual and textual information using a cross-modal attention mechanism.

In most of the above methods, the users' preferences are expressed by a few keywords \cite{10.1007/978-3-319-46484-8_1,8099712,10.1145/3123266.3123297,Zhang2018QCVS,9902992,10.1145/3323873.3325040,Xiao_Zhao_Zhang_Yan_Yang_2020,9063637,10.5555/3540261.3541333,10222138} or a short sentence \cite{10.5555/3504035.3504062,11298698,10.5555/3540261.3541333,10889812}. Contrary to these methods, SD-MVSum gets as input a long-form textual description of the desired video summary (the so-called script), thus allowing the generation of visually and semantically diverse summaries. Moreover, differently from \cite{10.1145/3746027.3755821} that considers just the visual content of the video, SD-MVSum leverages also the video's spoken content to discover further associations between the script and the video, and produce summaries that are better aligned with the content outlined in the script.

\subsection{Multimodal video summarization}

\begin{table*}[t]
\caption{Overview of large-scale datasets for generic (top three) and script-driven (bottom three) video summarization in the literature.}
\label{tab:datasets}
\resizebox{\textwidth}{!}{%
\begin{tabular}{|l|c|c|c|c|c|c|}
\hline
Dataset                                                         & Domains     & Videos & Data modalities                                                                     & \begin{tabular}[c]{@{}c@{}}Annotations\\ per video\end{tabular} & Type of annotations                                                                                        & Task                                                                                      \\ \hline \hline
\begin{tabular}[c]{@{}l@{}}VideoXum \cite{10334011}\\ \textcolor{gray}{(TMM'23)}\end{tabular}     & open domain & 14,001  & \begin{tabular}[c]{@{}c@{}}video, text\\ (video description)\end{tabular}           & 10                                                               & \begin{tabular}[c]{@{}c@{}}ground-truth video summaries,\\ text description of the video\end{tabular}      & \begin{tabular}[c]{@{}c@{}}video summarization\\ with multimodal output\end{tabular}      \\ \hline
\begin{tabular}[c]{@{}l@{}}MrHiSum \cite{10.5555/3666122.3667886}\\ \textcolor{gray}{(NeurIPS'23)}\end{tabular}  & 3,509       & 31,892  & video                                                                               & 1                                                                & frame-level importance scores                                                                              & \begin{tabular}[c]{@{}c@{}}video summarization \&\\ highlight detection\end{tabular}      \\ \hline
\begin{tabular}[c]{@{}l@{}}MMSum \cite{10655119}\\ \textcolor{gray}{(CVPR'24)}\end{tabular}       & 17          & 5,100   & \begin{tabular}[c]{@{}c@{}}video, text, transcripts\\ video metadata\end{tabular}   & 1                                                                & \begin{tabular}[c]{@{}c@{}}ground-truth video and text\\ summary\end{tabular}                              & \begin{tabular}[c]{@{}c@{}}multimodal summarization\\ with multimodal output\end{tabular} \\ \hline
\begin{tabular}[c]{@{}l@{}}S-VideoXum \cite{10.1145/3746027.3755821}\\ \textcolor{gray}{(ACM MM'25)}\end{tabular} & open domain & 11,908  & \begin{tabular}[c]{@{}c@{}}video, text \\(summary script)\end{tabular} & 10                                                               & \begin{tabular}[c]{@{}c@{}}ground-truth video summaries,\\ text descriptions of the summaries\end{tabular} & \begin{tabular}[c]{@{}c@{}}script-driven\\ video summarization\end{tabular}    \\ \hline \hline 
\begin{tabular}[c]{@{}l@{}}SM-VideoXum\\ (Ours)\end{tabular} & open domain & 11,908  & \begin{tabular}[c]{@{}c@{}}video, text (summary\\ script), transcripts\end{tabular} & 10                                                               & \begin{tabular}[c]{@{}c@{}}ground-truth video summaries,\\ text descriptions of the summaries\end{tabular} & \begin{tabular}[c]{@{}c@{}}script-driven multimodal\\ video summarization\end{tabular}    \\ \hline
\begin{tabular}[c]{@{}l@{}}SM-MrHiSum\\  (Ours)\end{tabular}      & 3,509       & 29,917  & \begin{tabular}[c]{@{}c@{}}video, text (summary\\ script), transcripts\end{tabular} & 1                                                                & \begin{tabular}[c]{@{}c@{}}ground-truth video summaries,\\ text descriptions of the summaries\end{tabular} & \begin{tabular}[c]{@{}c@{}}script-driven multimodal\\ video summarization\end{tabular}    \\ \hline
\end{tabular}}
\end{table*}

Several attempts have been made to advance the quality of automated video summarization using additional data modalities. Narasimhan et al. \cite{10.5555/3540261.3541333} examined the performance of CLIP-It when the textual input is formed as a set of auto-generated dense captions of the video content. Following, focusing on the summarization of instructional videos, Narasimhan et al. \cite{10.1007/978-3-031-19830-4_31} developed a method that takes into account the video frames and the audio transcripts, and selects video fragments showing important steps of the procedure; there are the fragments that are both most relevant to the task and also mentioned in the transcripts. Working also with instructional videos, Palaskar et al. \cite{palaskar-etal-2019-multimodal} performed video summarization using a multi-source sequence-to-sequence model with hierarchical attention, while a similar approach was adopted by Sanabria et al. \cite{9413097} for summarizing sports videos. Zhong et al. \cite{10.1145/3477538} built a method that creates semantically representative video summaries by minimizing the distance of learnable representations of the video content and its textual description, in a common embedding space. Argaw et al. \cite{10656029} presented a method that employs the visual content and a long-form description of it or the audio transcripts, and trained it with ground-truth pseudo-summaries obtained by prompting a Large Language Model (LLM) to extract the most informative moments from ASR transcripts. 

Furthermore, there are a few multimodal summarization approaches that aim to generate both visual and textual summaries. In this context, Fu et al. \cite{fu-etal-2021-mm} presented a method that integrates a jump-attention mechanism to align features extracted from transcripts and video frames, and trained it using multi-task learning to simultaneously optimize text and video summarization. He et al. \cite{10204014} built the A2Summ method, which aligns and attends multimodal inputs leveraging time correspondence using an alignment-guided self-attention mechanism; the latter learns how to form a keyframe-based and a text-based summary with the help of dual contrastive losses. Qui et al. \cite{10655119} used a hierarchy of cross-modal attention mechanisms to fuse visual features from video frames/fragments with textual features from audio transcripts, and create a video and a text summary using a frame scorer and a text encoder, respectively.

Finally, there is an increasing interest on leveraging the understanding and reasoning capabilities of modern LLMs for tackling the task of video summarization. Most such works follow a multi-step approach that begins by translating the video frames into a sequence of captions using a Multimodal LLM (e.g. LLaVA-1.5-7B \cite{NEURIPS2023_6dcf277e}). Then, some methods assess the importance of each frame using an LLM that takes into account the generated captions and the local and global context of the video \cite{11093752}, or knowledge obtained from ground-truth annotations about the frames' importance \cite{Sugihara2026}. Alternatively, some approaches use the sequence of generated captions to synthesize a textual summary using an LLM, and then select the video frames for inclusion in the summary by measuring the semantic distance between the frames' captions and the textual summary \cite{10.1145/3696409.3700273, 10.1007/978-981-96-5815-2_6}. Following a different direction, \cite{10.1609/aaai.v39i4.32374} trains an LLM-based method for video-to-video and video-to-text summarization using interleaved video frames and natural language temporal prompts. Besides their use as video summarizers, LLMs have been also employed for data augmentation, creating pseudo-ground-truth annotations to train smaller video summarization models \cite{10656029,Islam_2024_CVPR}.

The methods presented above produce generic summaries that aim to provide a complete synopsis of the entire video, thus not being tailored to specific needs about the summaries' content. Contrary to these methods, SD-MVSum takes into account such needs through a script that describes the desired content of the summary.

\begin{figure*}[t]
\centering
\includegraphics[width=0.95\textwidth]{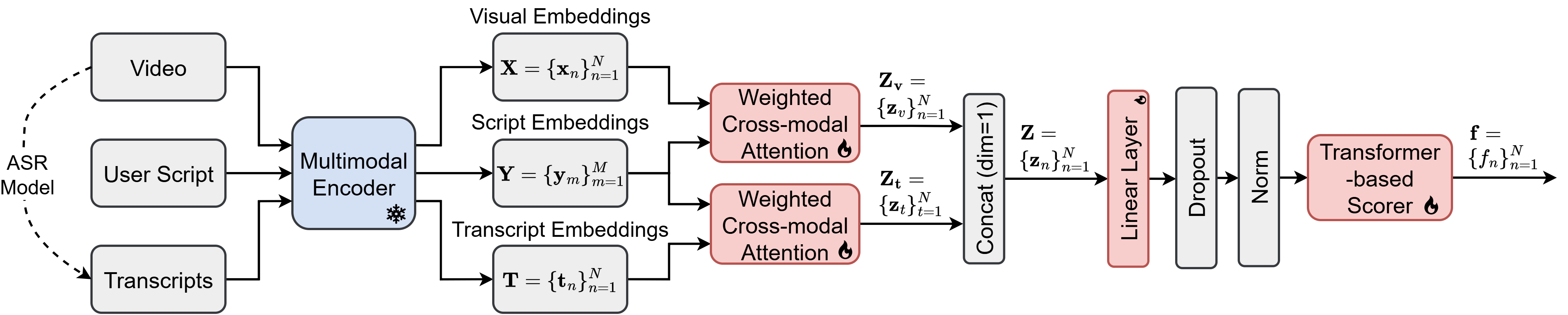}
\caption{Overview of the SD-MVSum network architecture. Given an input video, a script outlining the desired content of the summary, and a set of audio transcripts, SD-MVSum produces a video summary by finding associations of the script with both the visual and the spoken content in the video, using two weighted cross-modal attention mechanisms. The outputs of these mechanisms are concatenated and forwarded to a trainable Transformer-based scorer which computes frame-level importance scores. These scores are used by a frame/fragment selection component that forms the video summary given a video fragmentation and a time-budget about the summary duration.}
\label{fig:architecture}
\end{figure*}

\subsection{Video Summarization Datasets}

As discussed in \cite{10.1145/3746027.3755821}, most of the existing datasets for text/query-driven video summarization are either very small and cover a restricted set of domains (UT Egocentric \cite{10.1007/978-3-319-46484-8_1}, TV Episodes \cite{Yeung2014VideoSETVS}, QFVS \cite{8099712}, SumMe \cite{10.1007/978-3-319-10584-0_33}, TVSum \cite{7299154}, ARS \cite{10.1007/s11042-022-12442-w}), or contain annotations based on a small set of short-form (one/two-word) queries (RAD \cite{10.1145/3123266.3123297}). To tackle data scarcity, a few large-scale datasets for video summarization have been introduced in the literature over the last years. For example, the VideoXum dataset for cross-modal video summarization \cite{10334011} comprises $14,001$ open-domain videos up to $12.5$ min. long ($2$ min. avg. duration) with diverse visual content, derived from the ActivityNet Captions dataset \cite{8237345}. Each video is accompanied by $10$ ground-truth video summaries (in the form of binary frame-level annotations indicating the inclusion (label $1$) or not (label $0$) of a frame in the summary), obtained by $40$ different human annotators; and, a set of dense video captions that provide a high-level description of the full-length video. The MrHiSum dataset for video highlight detection and summarization \cite{10.5555/3666122.3667886} includes $31,892$ videos up to $5$ min. long ($3.3$ min. avg. duration), derived from the YouTube-8M dataset \cite{DBLP:journals/corr/Abu-El-HaijaKLN16}. Each video is associated with a series of frame-level importance scores (the so-called highlight labels in \cite{10.5555/3666122.3667886}) that have been computed after aggregating the viewing preferences of at least $50,000$ viewers of the video on YouTube, and used to formulate the ground-truth video summary based on the Knapsack algorithm and a predefined time-budget about the summary duration. The MMSum dataset for multimodal summarization and thumbnail generation \cite{10655119}, contains $5,100$ videos up to $115$ min. long ($14.5$ min. avg. duration), showing various everyday activities from $17$ main categories (e.g., cooking, sports, hobbies, travel). Each full-length video is accompanied by a ground-truth video and textual summary, as well as other metadata, such as title, author and category. Other large-scale datasets in the literature, such as MMS \cite{li-etal-2017-multi}, MSMO \cite{zhu-etal-2018-msmo}, How2 \cite{sanabria_02431947}, VMSMO \cite{li-etal-2020-vmsmo} and MM-AVS \cite{fu-etal-2021-mm}, contain ground-truth annotations that are suitable for training/evaluating methods that generate only textual summaries of the original video, and thus are out of the scope of this work. As shown in Table \ref{tab:datasets}, none of the datasets mentioned above provides the necessary data for training and evaluating script-driven video summarization methods. The only existing large-scale dataset for this task is S-VideoXum \cite{10.1145/3746027.3755821}, an extension of VideoXum which contains $11,908$ videos and $10$ different ground-truth summaries (binary frame-level annotations) and summary descriptions (the so-called ``scripts'' in \cite{10.1145/3746027.3755821}) per video. The available triplets of ``video, summary, and summary description'' can be used for training methods to produce different summaries for a given video, driven by the descriptions of the desired content of each summary.

In this work, we extend the S-VideoXum and MrHiSum datasets by generating and publicly releasing textual descriptions of the human-annotated summaries (for MrHiSum; updated such scripts are also produced for S-VideoXum) and audio transcripts (for both S-VideoXum and MrHiSum). In this way, we make these extended datasets, called SM-VideoXum and SM-MrHiSum, suitable for training and evaluating script-driven multimodal video summarization methods that take into account both the visual and the spoken content of the video. In contrast to most pre-existing datasets that are suitable for generic or query-driven summarization and not for script-driven summarization, the newly created datasets contain tenths (SM-MrHiSum) or hundreds (SM-VideoXum) of thousands of samples (i.e., ``video-script'' pairs) and the corresponding ground-truth summaries. These samples cover a wide range of video domains, allowing for the training and robust assessment of script-driven multimodal video summarization methods.

\section{Proposed SD-MVSum Method}
\label{sec:method}

\subsection{Problem statement}

Let us consider a full-length video and a script (composed of a number of sentences) outlining the content of the desired video summary. Different sentences of the script may refer to different parts of the full-length video with varying visual and semantic content. The goal of script-driven multimodal video summarization is to assess the relevance of the script with both the visual and the spoken content of the video, and select the video frames/fragments that are semantically associated to one or more sentences of the script and necessary for providing a complete synopsis of the video. The selected frames/fragments must form a concise video summary with a duration that is typically set to $15\%$ of the full-length video's duration \cite{9594911}.

\subsection{Network architecture}
\label{subsec:architecture}

\begin{figure*}[t]
\centering
\includegraphics[width=\textwidth]{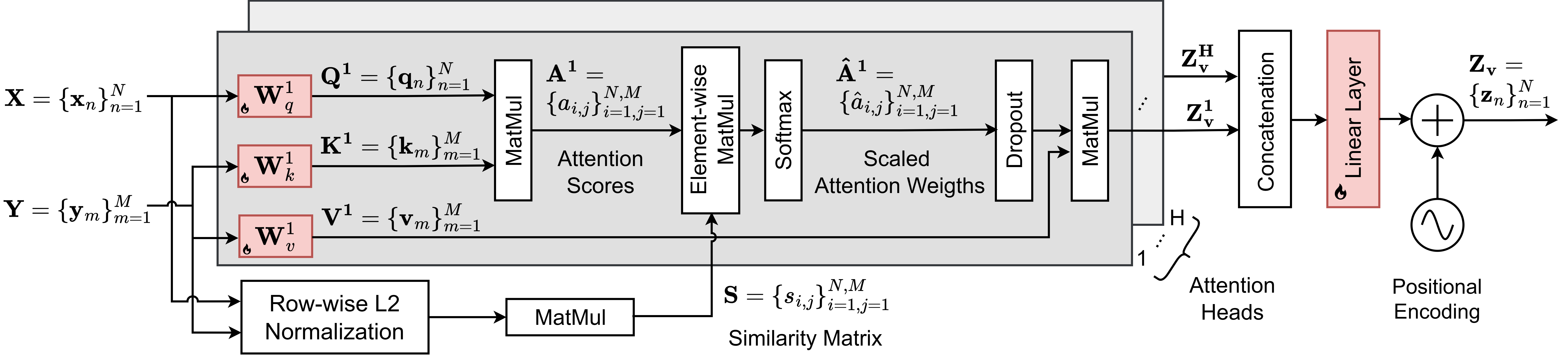}
\caption{The processing pipeline in the weighted cross-modal attention mechanism when fusing the visual and the script embeddings. The dynamic scaling of attention scores is performed based on the computed cosine similarity matrix $\mathbf{S}$ of the input embeddings.}
\label{fig:attention}
\end{figure*}

An overview of the SD-MVSum network architecture is provided in Fig. \ref{fig:architecture}. Let us assume a video of $N$ frames (after sampling one frame per second), a script outlining the content of the desired video summary formed by $M$ sentences, and a set of automatically extracted audio transcripts containing $K$ timestamped sentences. All these different input data pass through a pretrained multimodal encoder which produces three different sets of embeddings of the same size $D$; i.e., a set of visual embeddings ($\mathbf{X}=\{\mathbf{x}_n\}_{n=1}^N$), a set of script embeddings ($\mathbf{Y}=\{\mathbf{y}_m\}_{m=1}^M$), and a set of transcript embeddings ($\tilde{\mathbf{T}}=\{\mathbf{t}_k\}_{k=1}^K$). The transcript embeddings are expanded according to the timestamps of the associated transcripts, such that each embedding is repeated as many times as needed to match the number of video frames it spans, forming a new set $\mathbf{T} = \{\mathbf{t}_n\}_{n=1}^N$ that has the same number of embeddings with $\mathbf{X}$. This step is necessary for enabling, later on, the concatenation of the cross-modal embeddings that will come out of the weighted cross-modal attention mechanisms.

The script embeddings $\mathbf{Y}$ are fused with the visual embeddings $\mathbf{X}$ and transcript embeddings $\mathbf{T}$, via two weighted cross-modal attention mechanisms. The latter explicitly exploit the semantic similarity between a pair of data modalities, when modeling their dependence and forming the cross-modal embeddings. The concatenation of these embeddings ($\mathbf{Z_v} = \{{\mathbf{z}_v}\}_{n=1}^N$ and $\mathbf{Z_t} = \{\mathbf{z}_t\}_{t=1}^N$) is then performed, along the feature dimension, resulting in an overall set of cross-modal embeddings $\mathbf{Z} = \{\mathbf{z}_n\}_{n=1}^N$ with size $2D$, which are subsequently reduced in size by half, using a linear layer.

The embeddings obtained after dimensionality reduction pass through dropout and normalization layers, and are then given as input to a trainable Transformer-based scorer, which computes frame-level importance scores $\mathbf{f} = \{f_n\}_{n=1}^N$. These scores are finally used by a frame/fragment selection component that assembles the final summary, given a predefined temporal fragmentation of the full-length video and a fixed time-budget about the summary duration. As a note, despite the fact that SD-MVSum currently leverages the aforementioned data modalities, its design makes it easily extensible for taking into account additional modalities (e.g., dense video captions). Data from new modalities can be incorporated by introducing extra weighted cross-modal attention mechanisms and adjusting accordingly the input size of the employed linear layer for dimensionality reduction.

\subsection{Weighted cross-modal attention mechanism}
\label{wca}

The processing pipeline of fusing the visual and script embeddings, within the introduced weighted cross-modal attention mechanism, is depicted in Fig. \ref{fig:attention}. The same process, after replacing $\mathbf{X}$ with $\mathbf{T}$, is applied when fusing the script and transcript embeddings. So, given the $h^{th}$ attention head of the attention mechanism, the visual embeddings $\boldsymbol{X}$ pass through a linear layer of size $D/H$, where $H$ denotes the number of heads, forming the Query $\boldsymbol{Q^h} = \{\boldsymbol{q}_{n}\}_{n=1}^{N}$ matrix. The script embeddings $\boldsymbol{Y}$ pass through two different linear layers of size $D/H$, creating the Key $\boldsymbol{K^h} = \{\boldsymbol{k}_{m}\}_{m=1}^{M}$ and Value $\boldsymbol{V^h} = \{\boldsymbol{v}_{m}\}_{m=1}^{M}$ matrices. Then, the cross-modal embeddings in the output of each attention head, are computed as follows:
\begin{align*}
\mathbf{A^h} = \mathbf{Q^h}\mathbf{K^h}^T, \quad \mathbf{\hat{A}\mathbf{^h}} = \textnormal{Softmax}(\mathbf{A^h}\odot max(0,\mathbf{S})), \\ \mathbf{Z^h_v} = \mathbf{\hat{A}\mathbf{^h}} \mathbf{V^h} \quad \quad \quad \quad \quad \quad \quad \quad
\end{align*}
where, $\mathbf{K^h}^T$ is the transposed version of matrix $\mathbf{K^h}$ and $\mathbf{A^h}$ denotes the computed attention scores. $\mathbf{\hat{A}\mathbf{^h}}$ represents the scaled attention weights, computed by applying softmax to the results of an element-wise multiplication (denoted by $\odot$) between the attention scores and the zero-clipped matrix $\mathbf{S}$ ($max(0,\mathbf{S})$ denotes zero-clipping of negative values in $\mathbf{S}$). $\mathbf{S}$ is a cosine similarity matrix that is calculated as follows:
\begin{align*}
\mathbf{X_n} = \textnormal{L2}(\mathbf{X}), \quad \mathbf{Y_n} = \textnormal{L2}(\mathbf{Y}) \\ \mathbf{S} = \mathbf{X_n} \mathbf{Y_n}^T \quad \quad \quad
\end{align*}
with $\textnormal{L2}(\cdot)$ denoting L2 row-wise normalization. The output of the overall (multi-head) weighted cross-modal attention mechanism, is finally formulated as:
\begin{align*}
\mathbf{Z_v} = \textnormal{Concat}(\mathbf{Z_v^1}, \mathbf{Z_v^2}, \dots, \mathbf{Z_v^H}) \mathbf{W}^o + \textnormal{pe} ,
\end{align*}
where $\textnormal{pe}$ is the applied absolute positional encoding, that is implemented with the help of sine and cosine functions oscillating at different frequencies, as in \cite{10.5555/3295222.3295349}.

\begin{figure*}[t]
\centering
\includegraphics[width=\textwidth]{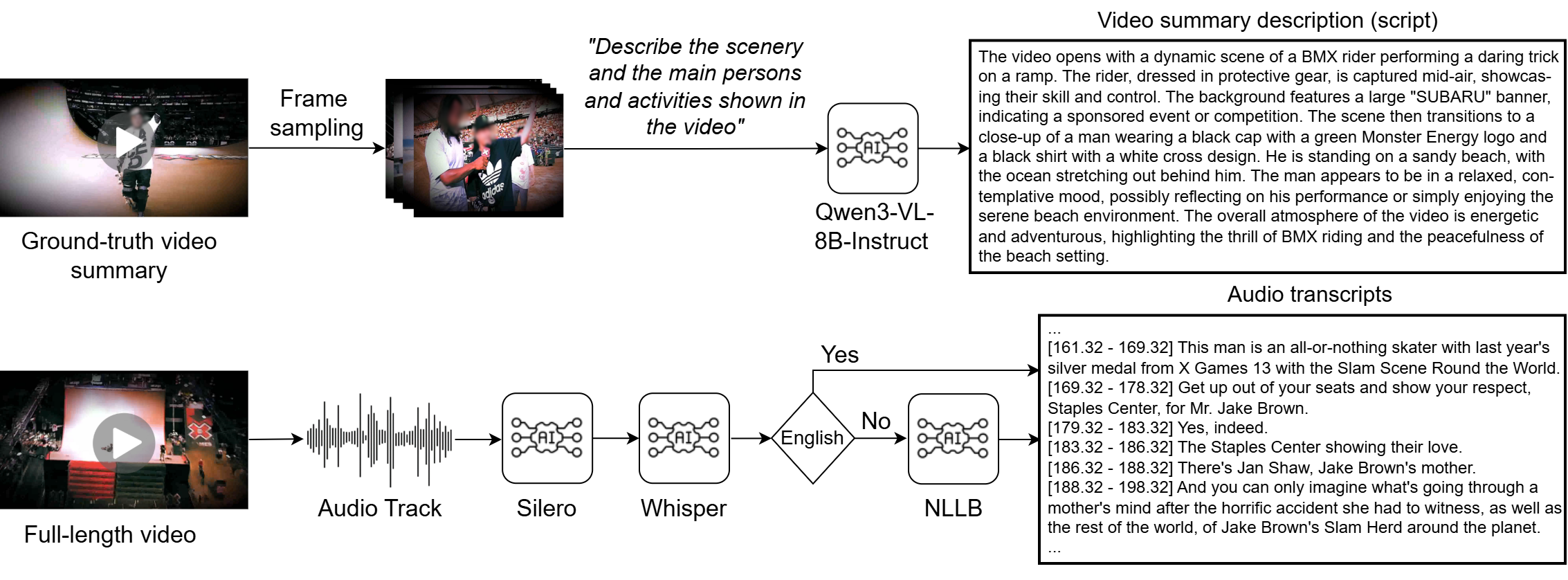}
\caption{Overview of the processing pipeline for creating the SM-VideoXum and SM-MrHiSum datasets for script-driven multimodal video summarization.}
\label{fig:dataset_creation}
\end{figure*}

So, instead of using a fixed scaling factor when computing the attention matrix (that is usually set equal to $1/\sqrt{d_k}$, where $d_k$ is the dimension of each individual attention head's Key/Query vectors, following \cite{10.5555/3295222.3295349}), the proposed weighted cross-modal attention mechanism performs a dynamic scaling of the computed attention scores using the similarity matrix $\mathbf{S}$ (see also Fig. \ref{fig:attention} for an illustration of this process). Theoretically, the expected range of values in $\mathbf{S}$ is $[-1, 1]$, where: $1$ indicates that the two embeddings point in the exact same direction, thus they are perfectly aligned; $0$ denotes that the two embeddings are orthogonal (perpendicular) to each other, and thus they have no linear relation; and $-1$ signifies that the two embeddings point in diametrically opposite directions. Nevertheless, in modern machine/deep learning, the computed scores for the embeddings obtained from many multimodal embedding models (e.g., CLIP) are usually positive (typically ranging between $0.1$ and $0.9$), because these models map data from different modalities into a shared semantic space where vectors share a general direction. This is validated experimentally, i.e., on the SM-VideoXum and SM-MrHiSum datasets. So, the introduced similarity matrix $\mathbf{S}$ acts as a ``suppression/preservation'' gate, where strongly-correlated elements in the common embedding space are preserved (without being amplified), while weakly-correlated ones are suppressed. In practice, if our method would encounter video-script pairs yielding negative cosine similarity values, it would set these values equal to zero, as indicated by the $max(0,\mathbf{S})$ function in the first block of equations of this section. This zero-clipping approach acts as a hard semantic filter; any negative relation in the common embedding space is driven to zero in the attention scores, which softmax converts into near-zero attention weight.

The effectiveness of this scaling is related to the following factors. i) Cross-modal attention between different modalities (like video pixels and written text) can be noisy. A standard Transformer calculates attention scores based on dot-products, which can easily be distracted by random, coincidental patterns across modalities; e.g., if a frame contains visual noise that happens to map highly to a script word, the computed attention score will be high. By calculating an explicit, semantic similarity matrix $\mathbf{S}$ bounded in $[0, 1]$ beforehand, we create a reliable de-noising filter, i.e., the computed attention score will be multiplied with a low similarity score for the corresponding frame-sentence pair, ensuring that the Transformer ultimately allocates attention only where a true semantic anchor exists. ii) Moreover, in a standard cross-attention mechanism, the model computes dot products across every frame-sentence pair and propagates gradients back through all of them. In long videos or videos with varying content, this causes the optimizer to waste significant effort on learning spurious, coincidental matches between completely unrelated text and visual features. The use of a semantic similarity matrix $\mathbf{S}$ bounded in $[0, 1]$ as a gate resolves this, by unifying forward-pass regularization with backward-pass gradient steering; i.e., it bounds the growth of attention scores and prevents overfitting to spurious visual-text correlations (regularization), while at the same time it vanishes the gradients for the irrelevant frame-sentence pairs and focuses backpropagation updates mainly on semantically relevant ones (gradient steering). iii) Finally, in video summarization where the goal is extreme compression (e.g., selecting only the top $15\%$ of relevant content), standard attention, by default, distributes small amounts of ``attention mass'' across all elements, creating a smooth, long-tailed distribution. Since the proposed approach suppresses the attention scores of weakly-correlated elements (typically, to near-zero), it forces the attention distribution to become highly sparse. For video summarization, this sparsity is an excellent inductive bias. It forces the network to ignore filler content, making it more effective in picking out the exact, distinct timestamps that match the provided script. Overall, the proposed dynamic scaling approach improves performance through a ``sculpting'' effect; by aggressively pruning away the irrelevant multimodal noise, the valuable correlations are the only things left standing.

\section{Proposed SM-VideoXum and SM-MrHiSum Datasets}

\subsection{Extended datasets construction}

The processing pipeline that was executed for constructing the extended SM-VideoXum and SM-MrHiSum datasets is presented in Fig. \ref{fig:dataset_creation}. As shown in the upper part of this figure, each human-annotated ground-truth summary is submitted to a frame sampling process that keeps one frame per second; the set of sampled frames is then given as input to a video-to-text component. To exploit the visual content understanding and description capacity of modern Multimodal Large Language Models, we replaced the LLaVA-NeXT-Video-7B model \cite{li2024llavanext-strong} that was used to this end in \cite{10.1145/3746027.3755821}, with the more powerful Qwen3-VL-8B-Instruct \cite{Qwen3-VL}. Moreover, based on experimentation with numerous prompts ($5$ in total) and through the qualitative analysis of the produced scripts, we replaced the prompt used to this end in \cite{10.1145/3746027.3755821} (\textit{``describe the important scenes in this video''}) with the following (more specific) one: \textit{``describe the scenery and the main persons and activities shown in the video.''} As before, the newly used Qwen3-VL-8B-Instruct model is prompted to generate a textual description of the ground-truth summary that is up to $200$ tokens long. The descriptions obtained for the ground-truth summaries of the SM-VideoXum and SM-MrHiSum datasets, through the aforementioned process, were used as input scripts in the evaluations reported in Section \ref{sec:experiments}.

\begin{figure*}[t]
\centering
\includegraphics[width=\textwidth]{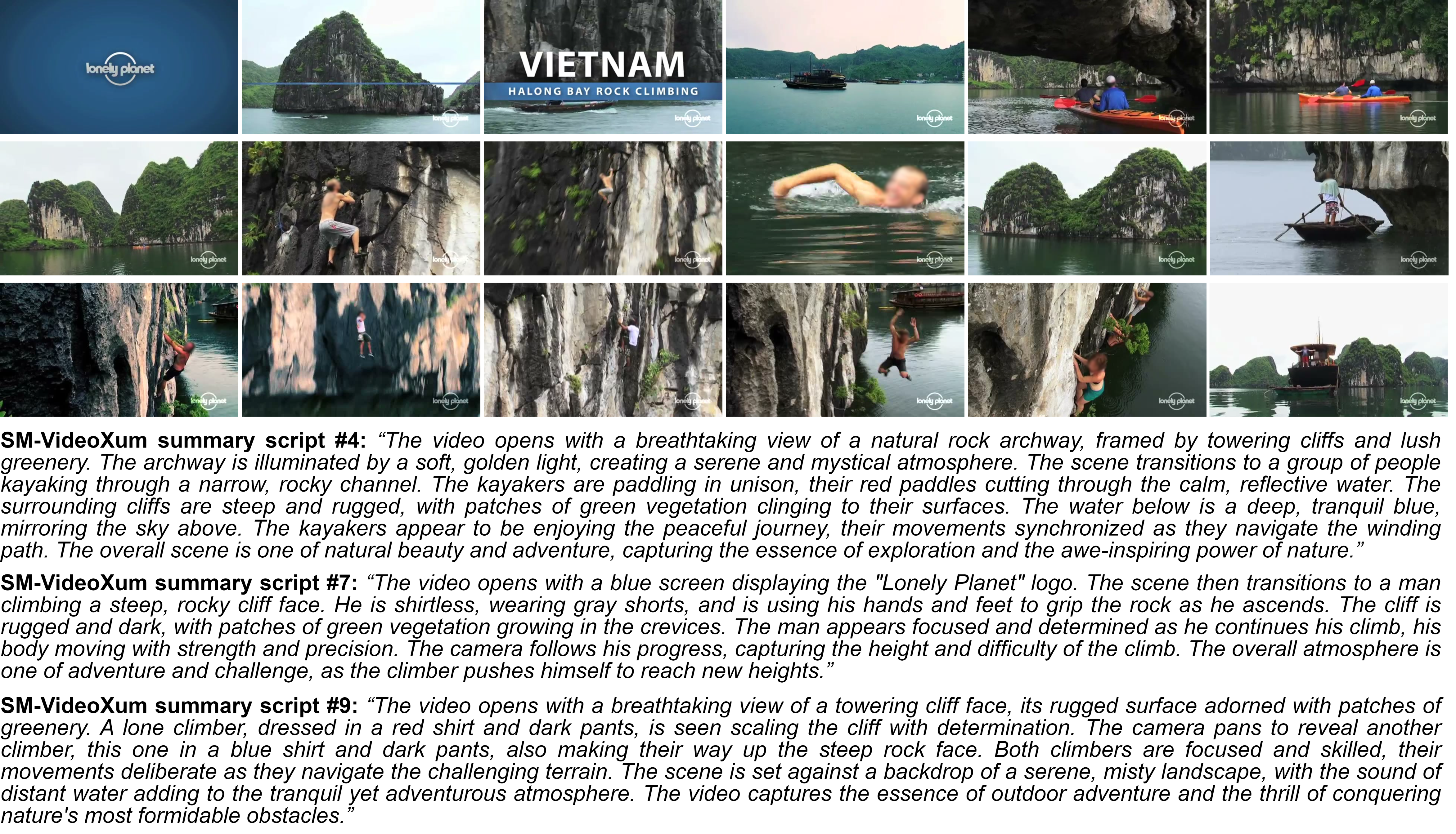}
\caption{A keyframe-based representation of the video ``v\_d3crFny-e3E'', and examples of the generated scripts for three ground-truth summaries of this video from the SM-VideoXum dataset.}
\label{fig:qual_example_a}
\end{figure*}

Focusing on the lower part of Fig. \ref{fig:dataset_creation}, each full-length video undergoes an audio transcript extraction process. For this, the audio stream of the video is submitted to a pretrained model of Silero VAD for voice activity detection \cite{SileroVAD}, which identifies the speech segments. The identified segments are then forwarded to a pretrained model of Whisper Turbo for speech-to-text transcription \cite{10.5555/3618408.3619590}, which outputs a set of timestamped transcripts. Finally, given that the employed multimodal encoder for obtaining embeddings from the input data has been trained on English textual data, any transcript in a different language is translated in English using the NLLB-200 model for machine translation \cite{nllbteam2022languageleftbehindscaling}. All the generated data and the full-length videos of the SM-MrHiSum and SM-VideoXum datasets, along with the extracted embeddings from visual and textual (script, transcript) data and the used data splits in our experiments, are publicly-available at: https://github.com/IDT-ITI/SD-MVSum.

\subsection{Dataset Quality Assessment}

To ensure the high quality of the constructed datasets, we started by carefully selecting the models that would be leveraged for this task, as described in the previous section. Qwen3-VL-8B-Instruct \cite{Qwen3-VL} was selected based on its substantially improved video understanding capabilities, compared to its ancestors (Qwen2.5-VL-7B and Qwen2-VL-7B), and its competitive performance in comparison to other SotA models (e.g., OpenAI GPT-5 nano \cite{singh2025openaigpt5card}) on several video question answer (e.g., RealWorldQA \cite{realworldqa2024}, MMStar \cite{10.5555/3737916.3738766}, SimpleVQA \cite{cheng2025simplevqa}) and video understanding benchmarks (e.g., Video-MME \cite{11093290}, MLVU \cite{11094860}, VideoMMMU \cite{hu2025videommmuevaluatingknowledgeacquisition}, MMVU \cite{11091892}). Silero VAD \cite{SileroVAD} offers very high accuracy and is robust to environmental noise, surpassing other popular open-source solutions, such as Google's WebRTC\footnote{T. Hanai, ``WebRTC based voice activity detection,'' [Online]. Available: https://github.com/reedom/VoiceActivityDetector. [Accessed: 30-Jul-2026]} \cite{mckinnon2026windowsizeversusaccuracy}. OpenAI's Whisper Turbo \cite{10.5555/3618408.3619590} is a SotA open model for multilingual automatic speech recognition, designed to balance high accuracy with significantly improved speed for speech-to-text transcription, that surpasses the performance of its ancestor WhisperX \cite{bain2022whisperx} and performs comparably with licensed solutions, such as Deepgram's Nova-2\footnote{Deepgram, ``Introducing Nova-2: The Fastest, Most Accurate Speech-to-Text API,'' [Online]. Available: https://deepgram.com/learn/nova-2-speech-to-text-api. [Accessed: 30-Jul-2026]}. It exhibits a low Word Error Rate (WER) on clean English datasets ($3-4\%$ WER on ``LibriSpeech Clean'' dataset) and a slightly larger batched WER (around $7.7\%$) on real-world datasets, while it achieves strong zero-shot cross-lingual performance after being trained on hundreds of thousands of hours on diverse audio across more than $100$ different languages. Meta AI's NLLB-200 model for multilingual machine translation \cite{nllbteam2022languageleftbehindscaling} is a SotA model with superior performance on the FLORES-200 dataset\footnote{N. Muennighoff, ``Flores200 Dataset Card,'' Dataset Documentation, Hugging Face, 2022. [Online]. Available: https://huggingface.co/datasets/Muennighoff/flores200. [Accessed: 30-Jul-2026]}, and performs well even for low-resource languages that were previously under-served by LLMs (e.g., GPT-4 \cite{openai2024gpt4technicalreport}, LLaMA-3 \cite{llama3modelcard}, Qwen-2.5 \cite{qwen2025qwen25technicalreport}).

\begin{table}[t]
\caption{Statistics about the scripts in the extended SM-VideoXum and SM-MrHiSum datasets.}
\label{tab:script_stats}
\begin{center}
\begin{tabular}{|lcccc|}
\hline
\multicolumn{1}{|l|}{}                 & \multicolumn{1}{c|}{Min} & \multicolumn{1}{c|}{Max} & \multicolumn{1}{c|}{Avg}   & Std  \\ \hline
\multicolumn{5}{|l|}{SM-VideoXum}                                                                                                \\ \hline
\multicolumn{1}{|l|}{\# words per script}     & \multicolumn{1}{c|}{41}  & \multicolumn{1}{c|}{187} & \multicolumn{1}{c|}{147.0} & 26.4 \\
\multicolumn{1}{|l|}{\# sentences per script} & \multicolumn{1}{c|}{3}   & \multicolumn{1}{c|}{25} & \multicolumn{1}{c|}{8.5}  & 1.8 \\ \hline
\multicolumn{5}{|l|}{SM-MrHiSum}                                                                                                 \\ \hline
\multicolumn{1}{|l|}{\# words per script}     & \multicolumn{1}{c|}{32}  & \multicolumn{1}{c|}{191} & \multicolumn{1}{c|}{148.0} & 25.4 \\
\multicolumn{1}{|l|}{\# sentences per script} & \multicolumn{1}{c|}{1}   & \multicolumn{1}{c|}{23} & \multicolumn{1}{c|}{8.6}  & 1.8 \\ \hline
\end{tabular}
\end{center}
\end{table}

\begin{table}[t]
\caption{Statistics about the audio transcripts in the extended SM-VideoXum and SM-MrHiSum datasets.}
\label{tab:speech_stats}
\begin{center}
\begin{tabular}{|lcccc|}
\hline
\multicolumn{5}{|c|}{Spoken content statistics}                                                                                                                                                           \\ \hline
\multicolumn{1}{|l|}{Dataset}              & \multicolumn{1}{c|}
{\begin{tabular}[c]{@{}c@{}}\# videos \\ total\end{tabular}} & \multicolumn{1}{c|}{\begin{tabular}[c]{@{}c@{}}\# videos \\  w/o speech\end{tabular}} & \multicolumn{2}{c|}{\begin{tabular}[c]{@{}c@{}}\% videos \\  w/o speech\end{tabular}}  \\ \hline
\multicolumn{1}{|l|}{SM-VideoXum} & \multicolumn{1}{c|}{11,908} & \multicolumn{1}{c|}{3,893}                                                       & \multicolumn{2}{c|}{32.7\%}      \\
\multicolumn{1}{|l|}{SM-MrHiSum}  & \multicolumn{1}{c|}{29,917} & \multicolumn{1}{c|}{5,638}                                                       & \multicolumn{2}{c|}{18.8\%}      \\ \hline
\multicolumn{5}{|c|}{For videos with
speech, \# sentences per transcript}                                                        \\ \hline
\multicolumn{1}{|l|}{Dataset}              & \multicolumn{1}{c|}{Min}    & \multicolumn{1}{c|}{Max}                                                         & \multicolumn{1}{c|}{Avg}  & Std  \\ \hline
\multicolumn{1}{|l|}{SM-VideoXum} & \multicolumn{1}{c|}{1}      & \multicolumn{1}{c|}{226}                                                         & \multicolumn{1}{c|}{21.6} & 20.1 \\
\multicolumn{1}{|l|}{SM-MrHiSum}  & \multicolumn{1}{c|}{4}      & \multicolumn{1}{c|}{310}                                                         & \multicolumn{1}{c|}{45.4} & 31.1 \\ \hline
\end{tabular}
\end{center}
\end{table}

To evaluate the quality of the generated scripts, we performed a visual inspection of the obtained scripts for $25$ randomly sampled videos from each of the SM-VideoXum and SM-MrHiSum datasets ($50$ videos in total). The selected videos included varying visual content from different categories, such as instructional videos (e.g., changing a tire, playing the violin, waxing a ski), TV shows and movies, music videos, and sports videos (skiing, climbing). Our evaluation indicated that the scripts are sensible, highly descriptive, can refer to multiple related or unrelated actions, may contain abstract cues and narrative elements, and correspond to the content of the relevant video summaries. These characteristics, that provide evidence about the factual correctness, consistency and semantic quality of the generated scripts, are illustrated in the example in Fig. \ref{fig:qual_example_a}; the first script relates to a video summary that focuses on the group of people kayaking, the second script relates to a video summary focusing on one of the rock climbers, and the third script relates to a video summary showing the efforts of two climbers and parts of the scenery. Moreover, the scripts in the SM-VideoXum and SM-MrHiSum datasets are diverse, varying from concise to extensive ones, as shown by the statistics (number of words and sentences per script) reported in Table \ref{tab:script_stats}. In the case of SM-VideoXum, we additionally calculated such statistics for each video across its ten available ground-truth summaries, and then averaged over the entire dataset. The averaged std of script words and sentences are $17.8$ and $1.3$, respectively, indicating that within-video diversity is also quite high.

To assess the quality of the ASR transcripts, we made a thorough listening test, taking into account the transcripts of the aforementioned $25$ randomly sampled videos from the SM-VideoXum dataset and computing the WER. As stated above, these videos included varying visual content from different categories and domains, while in most videos there was some background noise or music that made the task of automated transcription challenging. Our qualitative analysis documented the efficiency of the employed Whisper Turbo model. The computed WER score for this set of videos was equal to $0.082$, indicating the reasonably high faithfulness and quality of the ASR transcripts \cite{arif-etal-2025-wer}. Table \ref{tab:speech_stats} provides details about the number of videos without speech in the SM-VideoXum and SM-MrHiSum datasets, and statistics that document the  diversity of the obtained transcripts. As explained in Section \ref{subsec:details}, for videos without spoken content we use transcript embeddings with zero values.

Last but not least, the quality of the obtained English translations from other languages was not possible to be evaluated, since we have no access to human-annotated translated data for the videos of the SM-VideoXum and SM-MrHiSum datasets. Nevertheless, as stated in the beginning of this section, the employed NLLB-200 model achieves SotA performance on the FLORES-200 benchmarking dataset (exceeding previous translation SotA by an average of $44\%$), and performs well for low-resource languages.

\section{Experiments}
\label{sec:experiments}

\subsection{Evaluation protocol}
\label{subsec:eval_protocol}

We follow a slightly different evaluation approach on each dataset, to account for the differences in the available ground-truth annotations for each of them. 

For SM-VideoXum, based on the methodology in \cite{10334011}, we form the video summary by selecting the top-15\% scoring frames by the model, and quantify the similarity between the machine-generated and the ground-truth summary using the F-Score (\%). So, a given test video is matched with each one of the multiple available scripts for it, and each one of the generated summaries is compared with the corresponding ground-truth summary. Through this process, we compute an F-Score for each pair of compared summaries and we average these scores to form the final F-Score for this video. After performing this for all test videos of SM-VideoXum, we calculate the mean of the obtained F-Score values, forming a score that indicates the model's performance on the test set. 

\begin{table*}[t]
\caption{Performance comparisons with SotA methods for script-driven (upper part) and generic (lower part) video summarization on SM-VideoXum and SM-MrHiSum, in terms of F-Score (\%, denoted ``F1'') and Kendall's $\tau$ and Spearman's $\rho$ rank correlation coefficients. Best scores in bold, second-best scores underlined. Scores whose difference from those of SD-MVSum is not statistically significant are shown in italics.}
\label{tab:comparison}
\begin{center}
\begin{tabular}{|lc|ccc|ccc|ccc|}
\hline
\multicolumn{2}{|l|}{}                                                                                                          & \multicolumn{3}{c|}{Data modalities} & \multicolumn{3}{c|}{SM-VideoXum}                 & \multicolumn{3}{c|}{SM-MrHiSum}                  \\ \hline
\multicolumn{1}{|l|}{Task}                                                                              & Model                 & Script    & Visual    & Transcript   & F1            & $\tau$            & $\rho$            & F1            & $\tau$            & $\rho$            \\ \hline
\multicolumn{1}{|l|}{\multirow{3}{*}{\begin{tabular}[c]{@{}l@{}}Script-\\ driven\\ summ.\end{tabular}}} & SD-MVSum (proposed) & \checkmark       & \checkmark       & \checkmark          & \textbf{27.3} & N/A            & N/A            & \textbf{59.3} & \textbf{0.204} & {\ul 0.273}    \\
\multicolumn{1}{|l|}{}                                                                                  & SD-VSum \cite{10.1145/3746027.3755821} \textcolor{gray}{(ACM MM'25)}             & \checkmark       & \checkmark       & X           & {\ul 24.4}    & N/A            & N/A            & 57.9    & 0.175          & 0.236          \\
\multicolumn{1}{|l|}{}                                                                                  & CLIP-It \cite{10.5555/3540261.3541333} \textcolor{gray}{(NeurIPS'21)}              & \checkmark       & \checkmark       & X           & 22.8          & N/A            & N/A            & 56.3          & 0.120          & 0.169          \\ \hline
\multicolumn{1}{|l|}{\multirow{3}{*}{\begin{tabular}[c]{@{}l@{}}Generic\\ summ.\end{tabular}}}          & A2Summ \cite{10204014} \textcolor{gray}{(CVPR'23)}               & X        & \checkmark       & \checkmark          & 21.3          & 0.145          & 0.193          & {\ul 58.0}          & 0.169          & 0.239          \\
\multicolumn{1}{|l|}{}                                                                                  & CSTA \cite{10657498} \textcolor{gray}{(CVPR'24)}                 & X        & \checkmark       & X           & 23.5          & \textbf{0.176} & \textbf{0.233} & 57.7          & {\ul \textit{0.193}}    & \textit{\textbf{0.274}} \\
\multicolumn{1}{|l|}{}                                                                                  & PGL-SUM \cite{9666088} \textcolor{gray}{(IEEE ISM'21)}              & X        & \checkmark       & X           & 22.1          & {\ul 0.153}    & {\ul 0.203}    & 57.4          & 0.168          & 0.241          \\ \hline
\end{tabular}
\end{center}
\end{table*}

For SM-MrHiSum, we follow the evaluation strategy in \cite{10.5555/3666122.3667886} and formulate the video summary based on the machine-computed frame-level importance scores, a predefined temporal segmentation of the video, and a time-budget about the summary duration ($\leq 15\%$ of the video's length), by solving the Knapsack problem. Then, we quantify its similarity with the ground-truth summary using F-Score (\%) only once, since there is only one ground-truth summary per video. After performing this for all test videos of SM-MrHiSum, we average the obtained F-Score values, resulting in a score that indicates the model's performance on the test set. Moreover, since SM-MrHiSum contains ground-truth annotations also in the form of frame-level scores, we additionally apply the evaluation protocol of \cite{8954229}. Specifically, we quantify the alignment between the machine-computed and the ground-truth frame-level importance scores for a given video using the Kendall's $\tau$~\cite{kendall1945treatment} and Spearman's $\rho$~\cite{kokoska2000crc} rank correlation coefficients. The computed $\tau$ and $\rho$ values for all test videos are then averaged, defining the performance of the summarization model on the test set. When assessing the performance on generic video summarization, the same evaluation protocol of \cite{8954229} is applied also in the case of SM-VideoXum, after forming a single ground-truth summary per video by averaging its multiple binary ground-truth annotations at the frame-level. 

Both SM-MrHiSum and SM-VideoXum are divided into training, validation and test sets. During training, model selection is based on the validation set, i.e., by measuring the model's performance on the validation set after each training epoch. When training is completed, we keep the model with the highest validation-set performance, and assess it on the test set using the evaluation protocols described above.

\subsection{Implementation details}
\label{subsec:details}

Similarly to \cite{10334011} and \cite{10.5555/3666122.3667886}, videos are sampled at one frame per second, and embeddings (of size $D=512$) are obtained from the video frames, the script and the audio transcripts, using the CLIP model. In the case of videos without spoken content, we use transcript embeddings with zero values. For the samples of SM-VideoXum, we employ a fine-tuned CLIP model on the data of VideoXum, that has been released by the authors of \cite{10334011}\footnote{J. Lin, H. Hua, M. Chen et al., ``VideoXum GitHub repository,'' GitHub, 2023. [Online]. Available: https://videoxum.github.io/. [Accessed: 30-Jul-2026]}, while for the samples of SM-MrHiSum we use the CLIP ViT-B/32 model from HuggingFace\footnote{N. Reimers, P. Cuenca, T. Aarsen, ``clip-ViT-B-32 Model Card,'' Model documentation, Hugging Face, 2021. [Online]. Available: https://huggingface.co/sentence-transformers/clip-ViT-B-32. [Accessed: 30-Jul-2026]}. 

Each cross-modal attention mechanism of SD-MVSum contains $8$ heads. The frame scorer consists of a Transformer encoder, followed by a linear layer with $512$ neurons and a sigmoid activation to compute frame-level importance scores. The network's weights are initialized based on the Xavier uniform initialization approach (gain $= \sqrt{2}$, bias $=0.1$). Training on SM-VideoXum is based on the optimization of the BCE (Binary Cross-Entropy) loss between the predicted frame-level scores and the binary ground-truth labels, since this dataset does not include frame-level importance scores. Training on SM-MrHiSum is performed using the MSE (Mean Squared Error) loss, and the ground-truth frame-level importance scores. Training takes place for $50$ epochs in a batch mode with a batch size equal to $4$ and $64$ for SM-VideoXum and SM-MrHiSum, respectively, using the Adam optimizer and setting the learning rate, dropout rate and L2 regularization factor equal to $5\cdot10^{-5}$, $0.5$ and $10^{-4}$, respectively. All experiments were conducted on a workstation equipped with an Intel Core i5-11600K CPU and an NVIDIA RTX 3090 GPU. To allow the reproduction of our experiments, any used data and the PyTorch implementation of SD-MVSum have been made publicly-available at: https://github.com/IDT-ITI/SD-MVSum.

\subsection{Experimental comparisons and method ablations}
\label{subsec:comparisons}

\begin{table*}[t]
\centering
\caption{Performance comparison with variants of SD-MVSum on SM-VideoXum and SM-MrHiSum, in terms of F-Score (\%, denoted ``F1'') and Kendall's $\tau$ and Spearman's $\rho$ rank correlation coefficients. Best scores in bold. Scores whose difference from those of SD-MVSum is not statistically significant are shown in italics.}
\label{tab:ablation1}
\begin{tabular}{|lc|ccc|c|c|ccc|}
\hline
\multicolumn{2}{|l|}{}                                                                                                          & \multicolumn{3}{c|}{Data modalities} &         & SM-VideoXum & \multicolumn{3}{c|}{SM-MrHiSum}                  \\ \hline
\multicolumn{1}{|l|}{Task}                                                                              & Model                 & Script    & Visual    & Transcript   & Scaling & F1                 & F1            & $\tau$            & $\rho$            \\ \hline
\multicolumn{1}{|l|}{\multirow{3}{*}{\begin{tabular}[c]{@{}l@{}}Script-\\ driven\\ summ.\end{tabular}}} & SD-MVSum (proposed) & \checkmark       & \checkmark       & \checkmark          & \checkmark     & \textbf{27.3}      & \textbf{59.3} & \textbf{0.204} & \textbf{0.273} \\
\multicolumn{1}{|l|}{}                                                                                  & Variant \#1           & \checkmark       & \checkmark       & X           & \checkmark     & 26.6              & 58.3          & 0.169          & 0.230          \\
\multicolumn{1}{|l|}{}                                                                                  & Variant \#2           & \checkmark       & \checkmark       & \checkmark          & X      & 25.1              & 58.4          & \textit{0.195}          & \textit{0.265}          \\
\multicolumn{1}{|l|}{}                                                                                  & Variant \#3           & \checkmark       & \checkmark       & \checkmark          & convent.      & 24.9              & 58.5          & \textit{0.196}         & \textit{0.264}          \\ \hline
\end{tabular}
\end{table*}

We compared the proposed SD-MVSum method against a number of SotA methods for query/script-driven and generic (multimodal- or visual-based) video summarization. For the first class, we considered the SD-VSum \cite{10.1145/3746027.3755821} and CLIP-It \cite{10.5555/3540261.3541333} methods that were discussed in Section \ref{sec:literature}. For the second class, we took into account the A2Summ \cite{10204014} method for multimodal video summarization that also utilizes the audio transcripts, and two visual-based methods with SotA performance on video summarization benchmarks, namely the CSTA \cite{10657498} and PGL-SUM \cite{9666088} methods. 

The results of our evaluations are reported in Table \ref{tab:comparison}. The score differences between the proposed SD-MVSum method and the compared ones were tested in terms of statistical significance using a two-tailed paired T-Test with confidence level $\alpha = 0.05$, and were found to be statistically significant, with a few exceptions; for these exceptions, the corresponding compared method's score is shown in italics in Table \ref{tab:comparison}. As a note, this statistical significance testing is not applicable when comparing the proposed SD-MVSum with generic summarization methods on SM-VideoXum, due to the difference in the number of generated video summaries (generic summarization methods produce one tenth of the summaries that SD-MVSum and other script-driven methods produce, for this dataset). Nevertheless, these performance differences (SD-MVSum vs. generic summarization methods) are particularly pronounced, and also all considered generic summarization methods perform worse than SD-VSum, where the F-Score difference between SD-MVSum and SD-VSum is statistically significant. 

The comparison between script-driven video summarization methods in Table \ref{tab:comparison} showcases that SD-MVSum outperforms the other competing methods. This is attributed to the use of multiple embeddings for representing the script (as discussed in \cite{10.1145/3746027.3755821}), and algorithmic improvements over SD-VSum \cite{10.1145/3746027.3755821}, i.e., the dynamic scaling of the attention scores based on the computed cosine similarity matrix of the input embeddings, and the exploitation of the audio transcripts. These improvements led to statistically significant higher performance in both datasets, and according to all measures, compared to \cite{10.1145/3746027.3755821}, \cite{10.5555/3540261.3541333}. Further comparison with methods for generic summarization in Table \ref{tab:comparison} indicates the ability of SD-MVSum to produce video summaries that better reflect the target content outlined in the provided script. SD-MVSum outperforms all generic summarization methods on both datasets in terms of F-Score, and performs comparably with CSTA on SM-MrHiSum in terms of $\tau$ and $\rho$. 

To further stress the practical significance of the SD-MVSum's increased performance, we assessed the per-sample performance difference between SD-MVSum and SD-VSum. This assessment indicated that this difference in terms of F-Score exceeds $10$ percentage points for $27.5\%$ of the SM-VideoXum video-script pairs and $20.44\%$ of the SM-MrHiSum video-script pairs. This means that, for roughly one out of four or five video-script pairs, there is a particularly pronounced gain by applying SD-MVSum instead of the previous SD-VSum method.

As a remark, a cross-dataset evaluation of the trained models on SM-VideoXum and SM-MrHiSum (i.e., evaluating the performance of the SM-VideoXum-based trained model on SM-MrHiSum and vice versa) is not meaningful and thus was not conducted. As described in Section \ref{subsec:details}, these models are trained using different types of ground-truth annotations and training losses. So, despite the fact that both the original VideoXum and MrHiSum datasets (and as a consequence their SM-VideoXum and SM-MrHiSum extensions) have been introduced for the video summarization task, their ground-truth annotations tackle this task from a different perspective. Models trained on SM-VideoXum address the task by deciding whether a video frame should be part of the summary or not, while models trained on SM-MrHiSum address the task by estimating how important a video frame is, in relation to others.

Following, we conducted a series of ablation studies in order to examine: i) the contribution of each of the key concepts of SD-MVSum, namely, the use of audio transcripts as an auxiliary data source and the introduction of weighted cross-modal attention for modeling dependencies among different data modalities; ii) the impact of using fewer or more heads in the weighted cross-modal attention mechanisms; iii) the influence of smaller or larger batches of training data, iv) the use of other data fusion approaches for combining the output of the integrated weighted cross-modal attention mechanisms, v) the use of another data representation approach for the audio transcripts when there is no spoken content in the video (instead of using zero-valued vectors), and vi) the impact of taking into account also the visual-transcript relationship after modeling it via an extra cross-modal attention mechanism.

As before, the differences between SD-MVSum and its variants were tested in terms of statistical significance using the same approach, and were found to be statistically significant, with a few exceptions; for these exceptions, the corresponding variant's score is shown in italics. Moreover, we should stress that in Tables \ref{tab:ablation1}-\ref{tab:ablation6} reporting the results of these ablations, we always compare script-driven summarization approaches, thus for the SM-VideoXum dataset the evaluation protocol of \cite{8954229} involving the calculation of the Kendall's $\tau$ and Spearman's $\rho$ rank correlation coefficients is not applicable, as explained in Section \ref{subsec:eval_protocol}.

In our first ablation study, that aims to assess the contribution of the key concepts of SD-MVSum, we considered the following variants of SD-MVSum:
\begin{itemize}
    \item \textbf{Variant \#1} does not take into account the audio transcripts, and thus performs script-driven video summarization using only the visual content of the video.
    \item \textbf{Variant \#2} does not apply the proposed dynamic scaling of attention scores and follows a more straightforward data fusion approach, similarly to SD-VSum.
    \item \textbf{Variant \#3} again does not apply the proposed dynamic scaling of attention scores; instead, it applies conventional scaling by $1/\sqrt{d_k}$, as typically done in Transformers \cite{10.5555/3295222.3295349}.
\end{itemize}
The outcomes of this study, presented in Table \ref{tab:ablation1}, document the positive contribution of both of the aforementioned key concepts. More specifically, the removal of audio transcripts from the pool of input data (Variant \#1) leads to a consistent drop in the script-driven video summarization performance across both datasets and according to all measures (being more pronounced on SM-MrHiSum), pointing out the usefulness of audio transcripts when used as an auxiliary source of information. Moreover, using the computed attention scores in the cross-modal attention mechanisms ``as is'' (Variant \#2), or after scaling them by $1/\sqrt{d_k}$ (Variant \#3) leads to a measurable performance drop in terms of F-Score on SM-MrHiSum, i.e., close to $1\%$, and an even higher drop on SM-VideoXum, namely $>2\%$ (the observed small differences in $\tau$, $\rho$ were not found to be statistically significant). Such a finding demonstrates the strong contribution of the proposed weighted cross-modal attention mechanism in finding better dependencies among data from different modalities.

\begin{table}[t]
\centering
\caption{Performance of SD-MVSum on SM-VideoXum and SM-MrHiSum in terms of F-Score (\%) and Kendall's $\tau$ and Spearman's $\rho$ rank correlation coefficients, for different numbers of attention heads. Best scores in bold. The differences between the scores attained using the proposed number of attention heads and all other examined options are statistically significant.}
\label{tab:ablation2}
\begin{tabular}{|c|c|ccc|}
\hline
\multicolumn{1}{|l|}{} & SM-VideoXum    & \multicolumn{3}{c|}{SM-MrHiSum}                  \\ \hline
Att. heads             & F1            & F1            & $\tau$            & $\rho$            \\ \hline
4                      & 25.7          & 58.2          & 0.181          & 0.243          \\
8 (prop.)           & \textbf{27.3} & \textbf{59.3} & \textbf{0.204} & \textbf{0.273} \\
16                     & \textit{27.2}          & 58.4          & 0.185          & 0.251          \\ \hline
\end{tabular}
\end{table}

In our second ablation study, we examined the effect of using fewer or more attention heads in the weighted cross-modal attention mechanisms. In particular, starting from the number of heads in SD-VSum ($8$ in total), we measured the performance of SD-MVSum when using half and double of them. The results, reported in Table \ref{tab:ablation2}, indicate that the use of $8$-head weighted cross-modal attention mechanisms is the optimal choice for both datasets, and based on both of the employed evaluation protocols.

\begin{table}[t]
\centering
\caption{Performance of SD-MVSum on SM-VideoXum in terms of F-Score (\%), for different training data batch sizes. Best scores in bold. The differences between the scores attained using the proposed batch size and all other examined batch sizes are statistically significant.}
\label{tab:ablation3a}
\begin{tabular}{|c|c|}
\hline
\multicolumn{1}{|l|}{} & SM-VideoXum      \\ \hline
Batch size             & F1              \\ \hline
2                 & 26.7                 \\
4 (prop.)         & \textbf{27.3}        \\
8                 & 26.4                 \\ \hline
\end{tabular}
\end{table}

\begin{table}[t]
\centering
\caption{Performance of SD-MVSum on SM-MrHiSum in terms of F-Score (\%) and Kendall's $\tau$ and Spearman's $\rho$ rank correlation coefficients, for different training data batch sizes. Best scores in bold. Scores, whose difference from those attained using the proposed batch size is not statistically significant, are shown in italics.}
\label{tab:ablation3b}
\begin{tabular}{|c|ccc|}
\hline
\multicolumn{1}{|l|}{}    & \multicolumn{3}{c|}{SM-MrHiSum}            \\ \hline
Batch size          & F1            & $\tau$            & $\rho$      \\ \hline
32                  & 58.4          & 0.176          & 0.237          \\
64 (prop.)          & \textbf{59.3} & \textbf{0.204} & \textbf{0.273} \\
128                 &  58.7         &   \textit{0.195}    &      \textit{0.263}          \\ \hline
\end{tabular}
\end{table}

\begin{table}[t]
\centering
\caption{Performance of SD-MVSum on SM-VideoXum and SM-MrHiSum in terms of F-Score (\%) and Kendall's $\tau$ and Spearman's $\rho$ rank correlation coefficients, for different data fusion approaches. Best scores in bold. Scores, whose difference from those attained using the proposed fusion approach is not statistically significant, are shown in italics.}
\label{tab:ablation4}
\begin{tabular}{|l|c|ccc|}
\hline
                         & SM-VideoXum    & \multicolumn{3}{c|}{SM-MrHiSum}                  \\ \hline
Data fusion              & F1            & F1            & $\tau$            & $\rho$            \\ \hline
Mean pooling             & 26.4          &     58.5          &      0.182          &       0.247         \\
Max pooling              & 26.2          &      \textit{59.0}         &       0.184         &   0.248             \\
Concat. (prop.) & \textbf{27.3} & \textbf{59.3} & \textbf{0.204} & \textbf{0.273} \\ \hline
\end{tabular}
\end{table} 

In our third ablation study, we investigated the effect of using smaller or larger batches of training data. For SM-VideoXum, we started with a batch size equal to the one in \cite{10.1145/3746027.3755821} ($4$ videos, which means $40$ video-script pairs) and then we halved and doubled it. For SM-MrHiSum, we started with a similarly large batch size ($64$ video-script pairs), and we again halved and doubled it. The outcomes of this study, presented in Tables \ref{tab:ablation3a} and \ref{tab:ablation3b}, show that medium-sized batches of training data lead to the best performance on both datasets. 

In our fourth ablation study, we tried other approaches for fusing the embeddings in the output of the integrated cross-modal attention mechanisms. As alternatives to the proposed concatenation and the subsequent linear layer for dimensionality reduction (see Fig. \ref{fig:architecture}), we examined the use of mean pooling or max pooling. The results reported in Table \ref{tab:ablation4}, show that the employed data fusion approach is clearly more suitable compared to the tested alternatives, on both datasets.

In our fifth ablation study, we examined another approach for representing the audio transcripts in case there is no spoken content in the video. There are various approaches in the literature for representing input data from missing modalities when training multimodal network architectures, such as the use of vectors with constant values (all zeros or ones), the use of learned ``missing data'' embeddings, or the use of attention masking mechanisms. Nevertheless, the use of zero-valued embeddings appears to be one of the most common approaches \cite{9879085, bi-etal-2025-two, Ye2026}, and this is why we followed this approach in our work. In this way, the network learns to ignore this blank state and rely on the remaining data. As a note, the use of attention masking is not applicable in our case, since this approach can be used when a modality is partially missing, while in our case the audio transcripts are either completely available or entirely missing. So, in this ablation we evaluated an SD-MVSum variant that employs learned embeddings \cite{Ye2026} when there is no spoken content in the video. The results in Table \ref{tab:ablation5} document the competency of using zero-valued embeddings, in line with the dominant practice in the relevant literature.

\begin{table}[t]
\caption{Performance comparison with an SD-MVSum variant that employs learned embeddings when there is no spoken content (audio transcripts) in the videos of SM-VideoXum and SM-MrHiSum datasets, in terms of F-Score (\%, denoted “F1”) and Kendall's $\tau$ and Spearman's $\rho$ rank correlation coefficients. Best scores in bold.}
\label{tab:ablation5}
\resizebox{\columnwidth}{!}{%
\begin{tabular}{|l|c|ccc|}
\hline
\multirow{2}{*}{\begin{tabular}[c]{@{}l@{}}Missing data\\ representation\end{tabular}} & SM-VideoXum   & \multicolumn{3}{c|}{SM-MrHiSum}                 \\ \cline{2-5} 
                                                                                       & F1            & F1            & $\tau$            & $\rho$            \\ \hline
Zero vectors (prop.)                                                                   & \textbf{27.3} & \textbf{59.3} & \textbf{0.204} & \textbf{0.273} \\
Learned embeddings                                                                     & 26.0          & 58.6          & \textit{0.199} & \textit{0.268} \\ \hline
\end{tabular}}
\end{table}

\begin{table}[t]
\caption{Performance comparison with an extended variant of SD-MVSum that integrates a third weighted cross-modal attention mechanism to model the relationship between the visual content and the audio transcripts, on the SM-VideoXum and SM-MrHiSum datasets, in terms of F-Score (\%, denoted “F1”) and Kendall's $\tau$ and Spearman's $\rho$ rank correlation coefficients. Best scores in bold.}
\label{tab:ablation6}
\resizebox{\columnwidth}{!}{%
\begin{tabular}{|l|c|ccc|}
\hline
\multirow{2}{*}{} & SM-VideoXum   & \multicolumn{3}{c|}{SM-MrHiSum}                                   \\ \cline{2-5} 
                  & F1            & F1            & $\tau$                     & $\rho$                     \\ \hline
SD-MVSum (prop.)  & \textbf{27.3} & \textbf{59.3} & 0.204                   & 0.273                   \\
SD-MVSum$_{ext}$     & 26.7          & 58.7          & \textit{\textbf{0.207}} & \textit{\textbf{0.279}} \\ \hline
\end{tabular}}
\end{table}

In our sixth ablation study, we aimed to assess the influence of modeling also the relationship between the visual content and the audio transcripts. Intuitively, this should not contribute to better learning the task of script-driven video summarization, since the script outlining the desired content of the summary is not taken into account when examining such a relationship. However, to experimentally validate this, this relationship was modeled using a third weighted cross-modal attention mechanism that was introduced in SD-MVSum, where the visual embeddings were used as Query and Value and the transcript embeddings were used as Key. As shown in Table \ref{tab:ablation6}, the results of this ablation were mixed; compared to SD-MVSum, this extended variant of it achieved slightly higher $\tau$ and $\rho$ values but also slightly lower F-Score.

\subsection{Study on the impact of audio transcripts}

In this study we aimed to investigate if there are specific conditions where the use of audio transcripts has a more significant impact on the summarization performance. To do so, we divided the samples (``video-script'' pairs) of the test set of SM-VideoXum and SM-MrHiSum datasets in three different groups based on the correlation between the visual and the spoken content (low/medium/high correlation), and evaluated the method’s performance for each group. To quantify this correlation, we employed the computed cosine similarity matrix $\mathbf{S}$ for the visual and transcript embeddings and applied Dynamic Time Wrapping (DTW), given the fact that visual cues and spoken content may be slightly out of phase in some cases. To form the different groups of samples, i.e., to define the upper and lower boundary (correlation score) and count the number of samples for each group, we created a histogram of three equal-width bins using the computed correlation scores. The results of this evaluation, presented in Table \ref{tab:ablation7a}, reveal a clear relationship between the level of correlation of the visual and spoken content in a video, and the summarization performance; the larger the correlation, the higher the summarization performance in both datasets, according to all evaluation measures. This finding indicates that transcript information contributes most significantly to the video summarization performance, when there is a high correlation between the visual cues and the spoken content in the video. As a note, a low correlation between these data affects more significantly the summarization performance on the videos of the SM-MrHiSum dataset.

\begin{table}[t]
\caption{Performance evaluation of SD-MVSum on different subsets of samples (i.e., ``video-script'' pairs) of the SM-VideoXum and SM-MrHiSum test sets, that exhibit different levels of correlation (low, medium, high) between the visual content and the audio transcripts, in terms of F-Score (\%, denoted ``F1'') and Kendall's $\tau$ and Spearman's $\rho$ rank correlation coefficients. Best scores in bold.}
\label{tab:ablation7a}
\resizebox{\columnwidth}{!}{%
\begin{tabular}{|l|c|ccc|}
\hline
\multirow{2}{*}{\begin{tabular}[c]{@{}l@{}}Visual-transcript correl. \\ (\#samples per dataset)\end{tabular}} & SM-VideoXum   & \multicolumn{3}{c|}{SM-MrHiSum}                 \\ \cline{2-5} 
                                                                                                              & F1            & F1            & $\tau$            & $\rho$            \\ \hline
Low (3,490 / 35)                                                                                                & 26.8          & 55.1          & 0.070          & 0.101          \\
Medium (13,460 / 560)                                                                                           & 27.1          & 59.0          & 0.175          & 0.238          \\
High (6,120 / 956)                                                                                              & \textbf{28.0} & \textbf{59.7} & \textbf{0.232} & \textbf{0.307} \\ \hline
\end{tabular}}
\end{table}

\begin{table}[t]
\caption{Performance evaluation of SD-MVSum on different subsets of samples (i.e., ``video-script'' pairs) of the SM-VideoXum and SM-MrHiSum test sets, that exhibit different levels of correlation (low, medium, high) between the script and the audio transcripts, in terms of F-Score (\%, denoted ``F1'') and Kendall's $\tau$ and Spearman's $\rho$ rank correlation coefficients. Best scores in bold.}
\label{tab:ablation7b}
\resizebox{\columnwidth}{!}{%
\begin{tabular}{|l|c|ccc|}
\hline
\multirow{2}{*}{\begin{tabular}[c]{@{}l@{}}Script-transcript correl. \\ (\#samples per dataset)\end{tabular}} & SM-VideoXum   & \multicolumn{3}{c|}{SM-MrHiSum}                 \\ \cline{2-5} 
                     & F1            & F1            & $\tau$         & $\rho$         \\ \hline
Low (1,599 / 40)      & \textbf{27.7} & 54.3          & 0.023          & 0.036          \\
Medium (10,692 / 582)  & 27.1          & 59.2          & 0.206          & 0.278          \\
High (10,779 / 929)     & 27.4          & \textbf{59.7} & \textbf{0.216} & \textbf{0.288} \\ \hline
\end{tabular}}
\end{table}

\begin{table}[b]
\caption{Performance evaluation of SD-MVSum on two different subsets of samples (i.e., ``video-script'' pairs) of the SM-VideoXum and SM-MrHiSum test set, with and without spoken content, in terms of F-Score (\%, denoted ``F1'') and Kendall's $\tau$ and Spearman's $\rho$ rank correlation coefficients. Best scores in bold.}
\label{tab:ablation7c}
\resizebox{\columnwidth}{!}{%
\begin{tabular}{|l|c|ccc|}
\hline
\multirow{2}{*}{\begin{tabular}[c]{@{}l@{}}Transcript availability \\ (\#samples per dataset)\end{tabular}} & SM-VideoXum   & \multicolumn{3}{c|}{SM-MrHiSum}                 \\ \cline{2-5} 
                     & F1            & F1            & $\tau$         & $\rho$         \\ \hline
No (11,120 / 313)      & 27.2 & 58.7          & 0.185          & 0.250          \\
Yes (23,070 / 1,551)     & \textbf{27.3}          & \textbf{59.4} & \textbf{0.208} & \textbf{0.278} \\ \hline
\end{tabular}}
\end{table}

\begin{table*}[t]
\centering
\caption{Performance of SD-MVSum on SM-VideoXum and SM-MrHiSum datasets in terms of F-Score (\%) and Kendall's $\tau$ and Spearman's $\rho$ rank correlation coefficients, for different script-generation approaches. Best scores in bold. Scores, whose difference from those attained using the proposed script generation model and prompt is not statistically significant, are shown in italics.}
\label{tab:ablation8}
\begin{tabular}{|l|c|ccc|}
\hline
                                              & SM-VideoXum    & \multicolumn{3}{c|}{SM-MrHiSum}                  \\ \hline
Script-generation                             & F1            & F1            & $\tau$            & $\rho$             \\ \hline
LLaVA-Next-Video-7B \& prompt \#1             & 25.2          & \textit{58.8}          & 0.187          & 0.253          \\
Qwen3-VL-8B-Instruct \& prompt \#1            & 27.0          & \textit{58.9}          & \textit{0.199}          & \textit{0.269}          \\
Qwen3-VL-8B-Instruct \& prompt \#2 (proposed) & \textbf{27.3} & \textbf{59.3} & \textbf{0.204} & \textbf{0.273} \\ \hline
\end{tabular}
\end{table*}

Subsequently, we performed the same evaluation after grouping the samples of the test set of the SM-VideoXum and SM-MrHiSum datasets according to the correlation between the script and the spoken content in the video (low/medium/high correlation). In this case, we quantified this correlation using the computed cosine similarity matrix $\mathbf{S}$ for the script and transcript embeddings. Given the fact that there is no time correspondence between the script sentences and the audio transcripts, we calculated the global maximum cross-similarity (bag-of-concepts alignment) as follows:
\begin{align*}
  \textnormal{Cov}_{trans} &= \frac{1}{N}\sum\limits_{i=1}^{N}\max\limits_{1\leq j \leq M}(\mathbf{S}_{i,j}) \\
  \textnormal{Cov}_{script} &= \frac{1}{M}\sum\limits_{j=1}^{M}\max\limits_{1\leq i \leq N}(\mathbf{S}_{i,j}) \\
  \textnormal{Correl} &= \frac{\textnormal{Cov}_{trans} + \textnormal{Cov}_{script}}{2}
\end{align*}
where: transcript coverage (denoted $\textnormal{Cov}_{trans}$) finds the maximum similarity for every transcript sentence $i$ with any script sentence $j$, and averages these similarity scores across all $N$ sentences; script coverage (denoted $\textnormal{Cov}_{script}$) finds the maximum similarity for every script sentence $j$ with any transcript sentence $i$, and averages these similarity scores across all $M$ sentences; and the overall correlation (denoted $\textnormal{Correl}$) is computed as the arithmetic mean of the aforementioned coverage scores. As before, we formed the different groups of samples by mapping the computed correlation scores to a histogram of three equal-width bins. The outcomes of this evaluation, shown in Table \ref{tab:ablation7b}, indicate that the model trained on SM-VideoXum maintains similarly high levels of summarization performance in all cases, while the model trained on SM-MrHiSum performs well only when there is moderate or high correlation between the script and the spoken content. Similarly, the model trained on SM-MrHiSum performs poorly when there is weak correlation between the script and the audio transcripts.

Finally, to assess the impact of missing speech on the summarization performance, we divided the samples of the test set of each dataset into two groups, according to the existence of spoken content or not, and evaluated the performance of SD-MVSum on each of these groups. The results reported in Table \ref{tab:ablation7c}, show that the model trained on SM-VideoXum is not affected by the presence or absence of speech in the video, as it exhibits the same high levels of summarization performance on both groups of videos. Conversely, the model trained on SM-MrHiSum is positively influenced by the inclusion of spoken content; however, it maintains competitive performance even in the absence of speech, thereby demonstrating robustness in this case as well.

Overall, we observe that the video summarization performance depends on the quality of the audio transcripts, and their semantic alignment with both the visual content and the provided script. The audio transcript quality can be significantly affected by the performance of the model employed for ASR transcription and/or the existence of background noise in the video, as well as by missing or sparse speech. Hence, we acknowledge these factors as natural limitations of the proposed SD-MVSum method.

\subsection{Impact of dataset construction design choices}

Concerning dataset construction, we conducted an ablation study to investigate the impact of using scripts generated by different SotA Multimodal LLMs. More specifically, we trained and evaluated SD-MVSum using scripts obtained by:
\begin{itemize}
    \item Prompting LLaVA-NeXT-Video-7B to ``describe the important scenes in this video'' (prompt \#1), as in \cite{10.1145/3746027.3755821}.
    \item Prompting Qwen3-VL-8B-Instruct to ``describe the important scenes in this video'' (prompt \#1).
    \item Prompting Qwen3-VL-8B-Instruct to ``describe the scenery and the main persons and activities shown in the video'' (prompt \#2).
\end{itemize}
The findings are presented in Table \ref{tab:ablation8}. A comparison between the scripts obtained by LLaVA-Next-Video-7B and Qwen3-VL-8B-Instruct for the same prompt (prompt \#1), documents the competency of the latter model to create scripts that are more useful for learning the task of script-driven multimodal video summarization, a finding that is more pronounced on the videos of the SM-VideoXum dataset. In addition, prompting the Qwen3-VL-8B-Instruct model with a more detailed and specific prompt for generating these scripts (prompt \#2) leads to further measurable advancements in the summarization performance on both datasets, according to all measures. The observed differences on SM-MrHiSum after modifying the prompt to the Qwen3-VL-8B-Instruct model were not found to be statistically significant; however, the differences when using LLaVA-Next-Video-7B (as in \cite{10.1145/3746027.3755821}) for generating the scripts were statistically significant, according to the attained rank correlation coefficients. These findings highlight the added value of the released datasets.

\begin{figure*}[h!]
\centering
\includegraphics[width=0.99\textwidth]{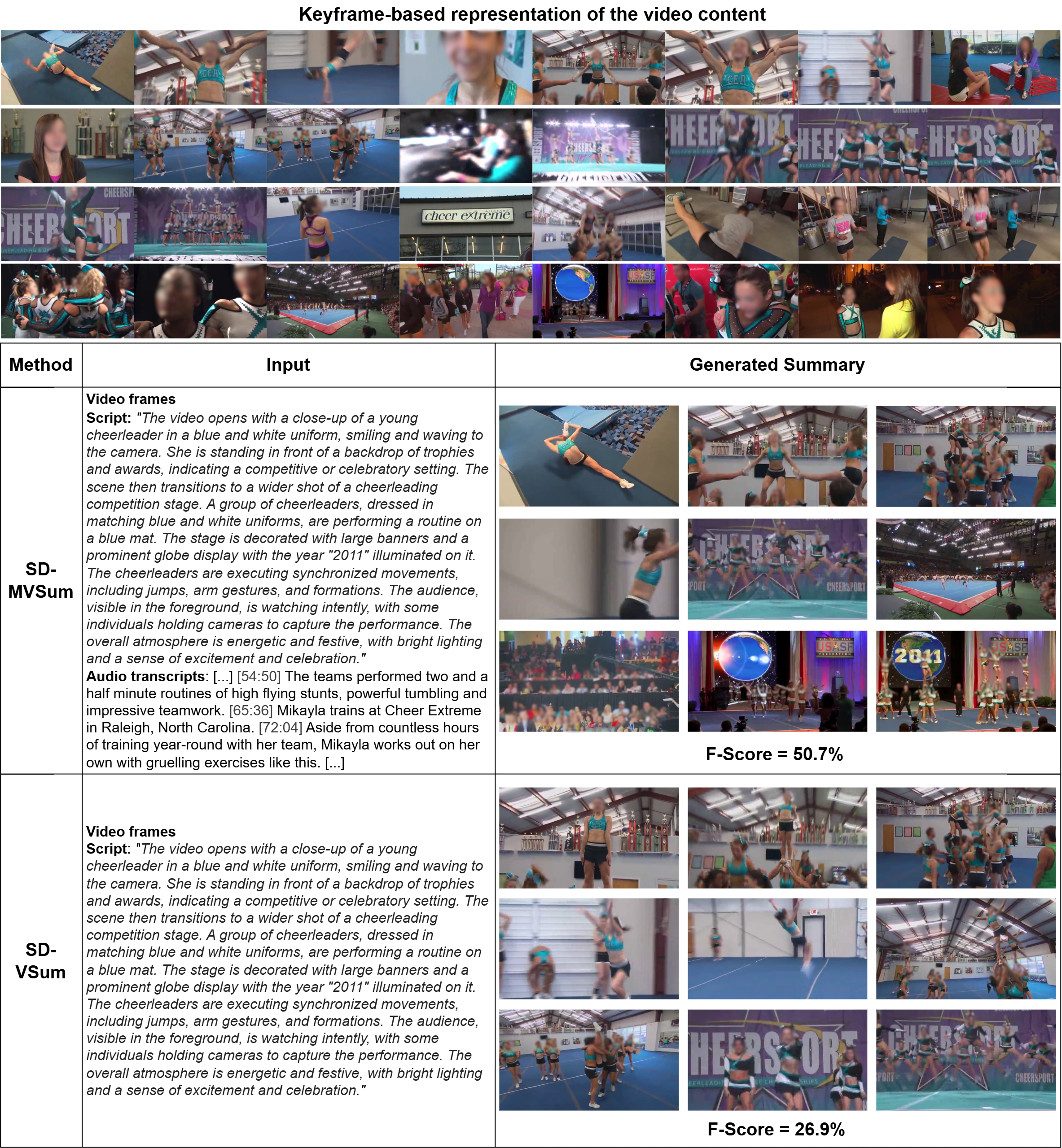}
\caption{An indicative example from our qualitative analysis. The upper part provides a keyframe-based representation of the content of the full-length video, and the tabular structure beneath shows the utilized input data and the generated video summary by each method.}
\label{fig:qual_example_b}
\end{figure*}

\begin{figure*}[t!]
\centering
\includegraphics[width=0.87\textwidth]{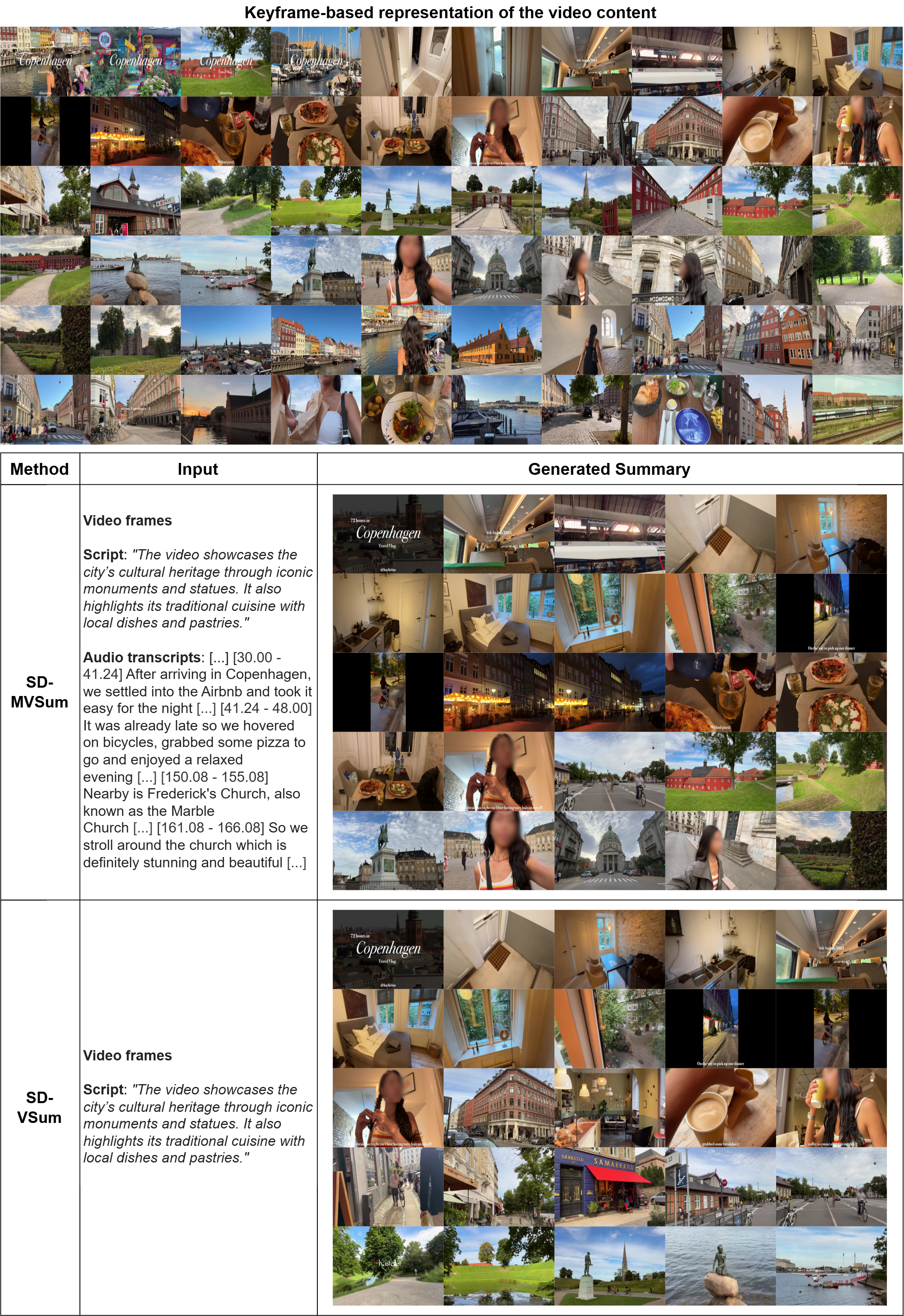}
\caption{An indicative example from our qualitative analysis. The upper part provides a keyframe-based representation of the content of the full-length video, and the tabular structure beneath shows the utilized input data and the generated video summary by the SD-MVSum and SD-VSum methods.}
\label{fig:qual_example_c}
\end{figure*}

\begin{figure*}[t!]
\centering
\includegraphics[width=0.97\textwidth]{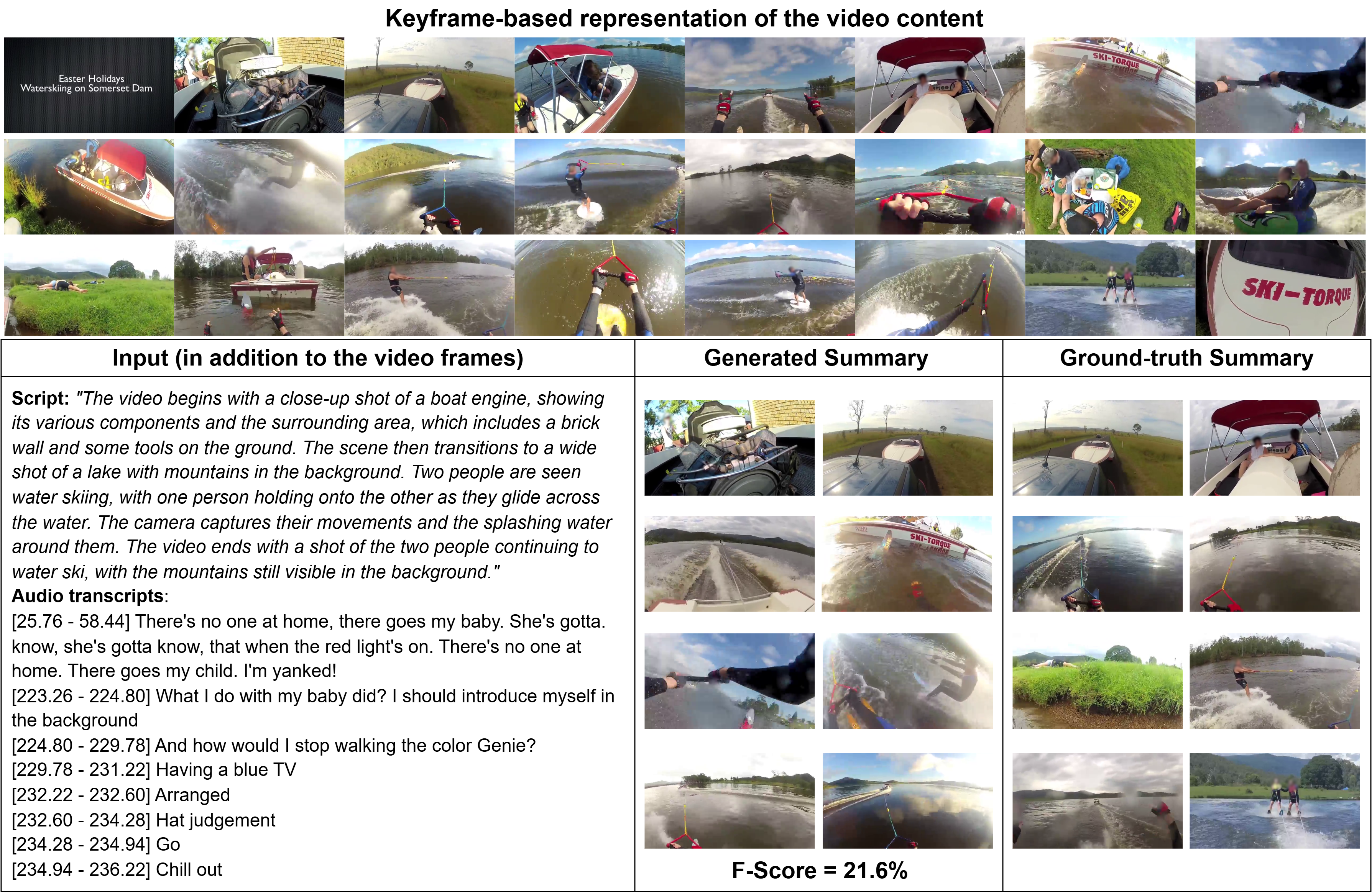}
\caption{An indicative sample from our qualitative analysis. The upper part provides a keyframe-based representation of the content of the full-length video, and the tabular structure beneath shows the utilized input data, the generated video summary by the proposed SD-MVSum method, and the ground-truth summary for this video.}
\label{fig:qual_example_d}
\end{figure*}

\subsection{Qualitative analysis}

To further evaluate the contribution of audio transcripts in the script-driven video summarization outcome, we performed a qualitative analysis that was based on manual observation of the generated summaries by our SD-MVSum method and the SD-VSum method \cite{10.1145/3746027.3755821} that uses just the visual content of the video, for a set of sampled videos from the SM-VideoXum and SM-MrHiSum datasets. One of the examined samples is presented in Fig. \ref{fig:qual_example_b}. The upper part provides a keyframe-based representation of the content of the full-length video, and the tabular structure beneath shows the utilized input data and the generated video summary by each method. As can be seen, both methods focused on parts of the video presenting cheer-leading routines (either during training or at a competition) and ignored less relevant parts showing e.g., interviews, thus being aligned with the input script. However, SD-VSum puts emphasis on parts of the video showing the group's training in an indoor area (choosing $7$ relevant video fragments) and focuses less on parts of the video presenting the team's participation at the competition (selecting $2$ relevant video fragments). On the contrary, SD-MVSum produces a more comprehensive summary, showing parts from the training process (including $4$ relevant video fragments), capturing the essence of the competition stage (using $2$ relevant fragments), and presenting the team's performance during the competition (selecting $3$ relevant fragments). This example, demonstrates that the use of audio transcripts allowed SD-MVSum to spot more effectively video parts showing the team's preparation and participation at a competition, and generate a summary that is more aligned with the viewer's needs, as indicated by the significantly higher F-Score.

Another example from our qualitative analysis using a travel VLOG from YouTube (available at: https://www. youtube.com/watch?v=F2RLoK16U4k) is presented in Fig. \ref{fig:qual_example_c}. The upper part provides a keyframe-based representation of the content of the full-length video, and the tabular structure beneath shows the utilized input data and the generated video summary by the SD-MVSum and SD-VSum methods. As can be seen, both methods provide a complete synopsis of the travel and focus on iconic places and monuments of the city. However, SD-MVSum pays more attention to video parts showing local dishes and pastries, producing a video summary that is more tailored to the input script.

The final example from our qualitative analysis is depicted in Fig. \ref{fig:qual_example_d}. As before, the upper part shows a keyframe-based synopsis of the visual content of the full-length video. The tabular structure beneath presents the employed input data by the proposed SD-MVSum method, along with the generated and the ground-truth summary for this video. As a note, the video shows a group of people waterskiing and having a picnic on the Somerset Dam. Over the entire video, there is some background music, namely a song (from the hard rock genre). The existence of this background music resulted in the extraction of audio transcripts that capture (in part) the song's lyrics, having limited semantic correlation with both the visual content and the input script. These distracted, to some extent, SD-MVSum, which produced a video summary that does not fully align with the script, as indicated by the low F-Score value ($21.6\%$). This finding is in line with our remarks about the effect of the audio transcripts in the video summarization performance, and the natural limitations of the proposed method.

\section{Conclusions}
In this paper, we presented the SD-MVSum method for script-driven multimodal video summarization, which takes into consideration the relevance of the provided script with both the visual and the spoken content in the video. This relevance is modeled using a new weighted cross-modal attention mechanism, which exploits the semantic similarity between paired modalities and applies a dynamic scaling to promote the video parts that are most relevant to the script. To assist the training and evaluation of script-driven multimodal video summarization methods, we extended two large-scale datasets for video summarization (S-VideoXum, MrHiSum) to make them suitable for the task. Our quantitative and qualitative evaluations showcased the competitiveness of SD-MVSum against other SotA methods for script-driven and generic video summarization.

{
    \small


}

\begin{IEEEbiography}
[{\includegraphics[width=1in,height=1.25in,clip,keepaspectratio]{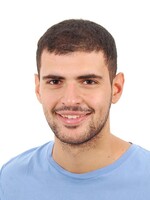}}]{Manolis Mylonas} received a Diploma degree from the Dept. of Electrical and Computer Engineering of the Aristotle University of Thessaloniki in July 2023. From February 2024 to September 2025 he worked as a Research Assistant at the Information Technologies Institute of CERTH, investigating deep leaning based methods for video summarization. He is currently a PhD candidate at the Center for Brain and Cognition of the Universitat Pompeu Fabra in Barcelona, Spain. His research interests cover decision making, reinforcement learning and multi-agent systems. 
\end{IEEEbiography}

\begin{IEEEbiography}[{\includegraphics[width=1in,height=1.25in,clip,keepaspectratio]{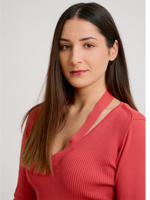}}]{Charalampia Zerva} holds a Bachelor's degree in Applied Informatics, specializing in Computer Science and Technology, from the University of Macedonia (2024). Since then, she works as a Research Assistant in the field of deep learning, at the Information Technologies Institute of CERTH. From October 2025, she is pursuing a Master’s degree in Digital Media and Computational Intelligence at the Aristotle University of Thessaloniki.
\end{IEEEbiography}

\begin{IEEEbiography}[{\includegraphics[width=1in,height=1.25in,clip,keepaspectratio]{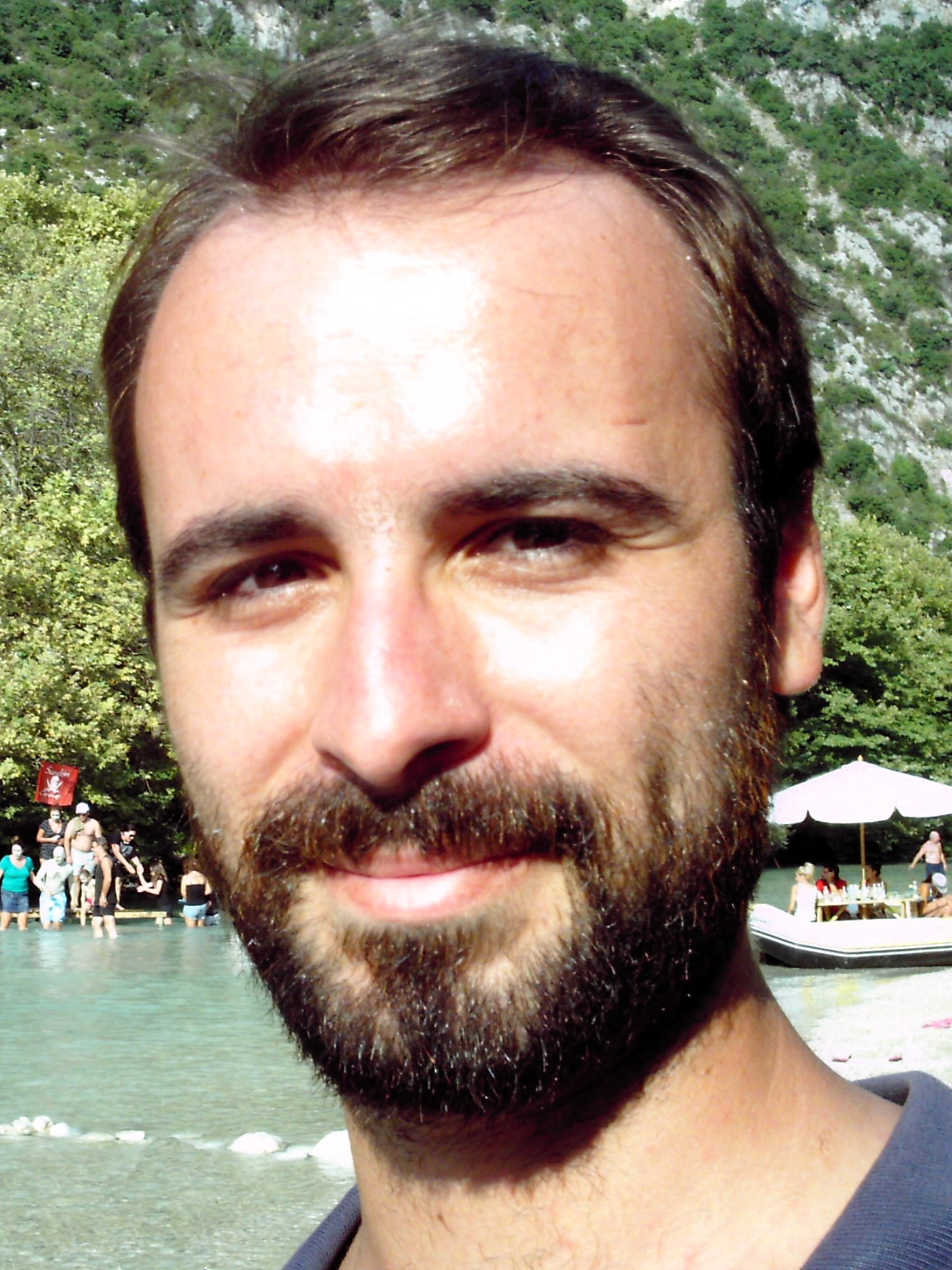}}]{Evlampios Apostolidis} received a Diploma degree from the Dept. of Electrical and Computer Engineer of the Aristotle University of Thessaloniki in July 2007, a Master degree in Information Systems from the University of Macedonia in November 2011, and a PhD degree in Electronic Engineering from the Queen Mary University of London in August 2023. From January 2012 he works as a Research Associate at the Information Technologies Institute of CERTH, in Thessaloniki, Greece. He has co-authored 6 journal papers, 5 book chapters and more than 40 conference papers. His research interests lie in the area of explainable AI and deep learning for video analysis and understanding, for applications in the Media sector.
\end{IEEEbiography}

\begin{IEEEbiography}[{\includegraphics[width=1in,height=1.25in,clip,keepaspectratio]{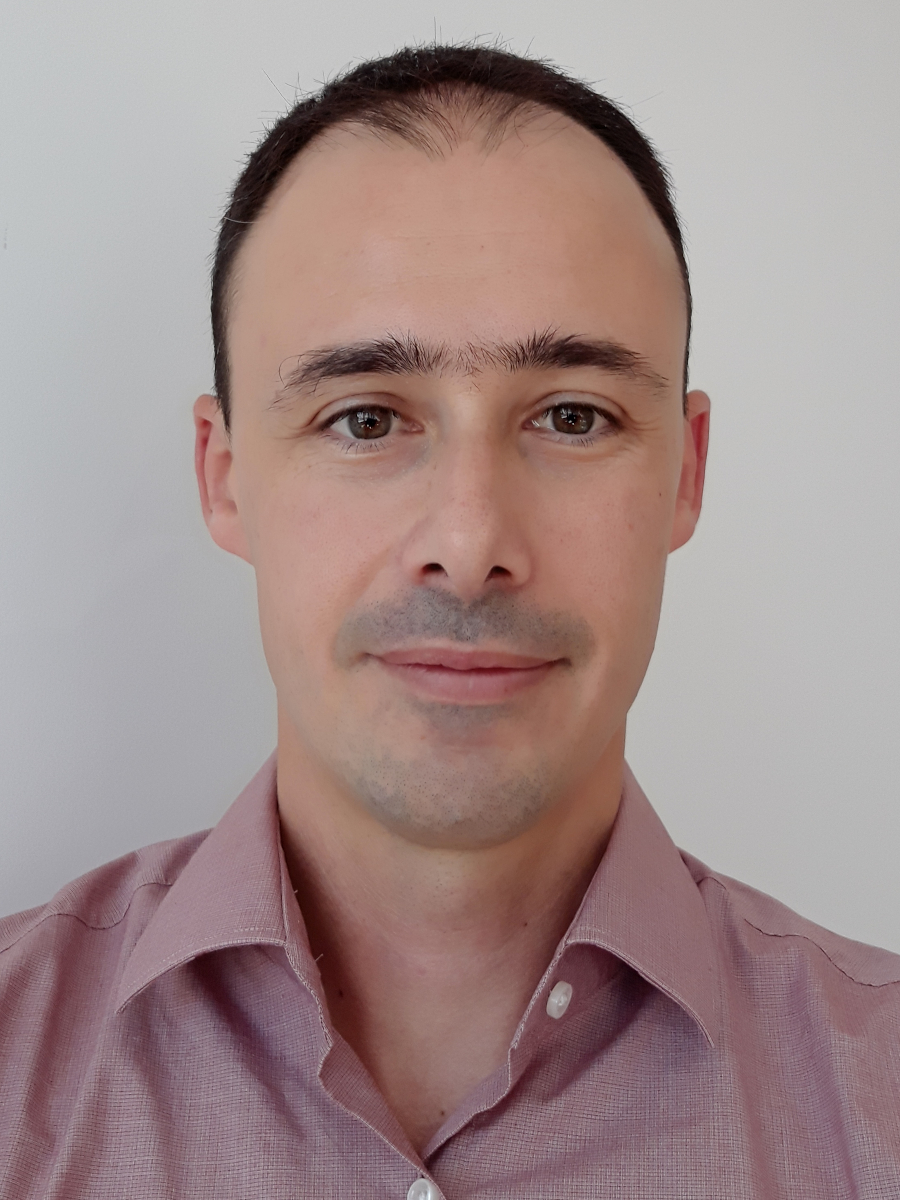}}]{VASILEIOS MEZARIS} (Senior Member, IEEE) received the B.Sc. and Ph.D. degrees in electrical and computer engineering from the Aristotle University of Thessaloniki, Greece, in 2001 and 2005, respectively. He is currently a Research Director of the Centre for Research and Technology Hellas, Information Technologies Institute. He has coauthored more than 50 journal articles, 20 book chapters, 200 conference papers, and three patents. His research interests include multimedia understanding and artificial intelligence; in particular, image and video analysis and annotation, machine learning and deep learning for multimedia understanding and big data analytics, explainable AI and green AI, multimedia indexing and retrieval, and applications of multimedia understanding and artificial intelligence. He served as Senior Area Editor for IEEE SIGNAL PROCESSING LETTERS, and as Associate Editor for IEEE SIGNAL PROCESSING LETTERS and IEEE TRANSACTIONS ON MULTIMEDIA.
\end{IEEEbiography}

\EOD

\end{document}